\documentclass[final]{cvpr}

\usepackage{times}
\usepackage{epsfig}
\usepackage{graphicx}
\usepackage{amsmath}
\usepackage{amssymb}
\usepackage{algorithm}
\usepackage{algpseudocode}
\usepackage{booktabs}
\usepackage{pifont}
\usepackage[numbers,sort&compress]{natbib}

\usepackage[pagebackref=true,breaklinks=true,colorlinks,bookmarks=false]{hyperref}

\def\cvprPaperID{5198} 
\def\confYear{CVPR 2021}

\newcommand{\Sref}[1]{Sec.~\ref{#1}}
\newcommand{\Eref}[1]{Eq.~(\ref{#1})}
\newcommand{\Fref}[1]{Fig.~\ref{#1}}
\newcommand{\Tref}[1]{Table~\ref{#1}}
\newcommand{\Aref}[1]{Algorithm~\ref{#1}}

\newcommand{\cellimgsmall}[1]{
    \includegraphics[width=2.0cm, height=2.0cm]{#1}
}
\newcommand{\cellimgtiny}[1]{
    \includegraphics[width=2.0cm, height=2.0cm]{#1}
}
\newcommand{\cellimgsixperrow}[1]{
    \includegraphics[width=2.3cm, height=2.3cm]{#1}
}
\newcommand{\minisection}[1]{\vspace{1mm}\noindent{\bf #1}}

\begin{document}

\title{Explore Image Deblurring via Encoded Blur Kernel Space}

\author{
Phong Tran$^{1}$ \quad Anh Tuan Tran$^{1,2}$ \quad Quynh Phung $^{1}$ \quad Minh Hoai$^{1,3}$ \\
$^1$VinAI Research, Hanoi, Vietnam,
$^2$VinUniversity, Hanoi, Vietnam,\\
$^3$Stony Brook University, Stony Brook, NY 11790, USA\\
{\tt\small \{v.phongtt15,v.anhtt152,v.quynhpt29,v.hoainm\}@vinai.io}
}

\maketitle

\def\mA{\mathcal{A}}
\def\mB{\mathcal{B}}
\def\mC{\mathcal{C}}
\def\mD{\mathcal{D}}
\def\mE{\mathcal{E}}
\def\mF{\mathcal{F}}
\def\mG{\mathcal{G}}
\def\mH{\mathcal{H}}
\def\mI{\mathcal{I}}
\def\mJ{\mathcal{J}}
\def\mK{\mathcal{K}}
\def\mL{\mathcal{L}}
\def\mM{\mathcal{M}}
\def\mN{\mathcal{N}}
\def\mO{\mathcal{O}}
\def\mP{\mathcal{P}}
\def\mQ{\mathcal{Q}}
\def\mR{\mathcal{R}}
\def\mS{\mathcal{S}}
\def\mT{\mathcal{T}}
\def\mU{\mathcal{U}}
\def\mV{\mathcal{V}}
\def\mW{\mathcal{W}}
\def\mX{\mathcal{X}}
\def\mY{\mathcal{Y}}
\def\mZ{\mathcal{Z}} 

\def\bbN{\mathbb{N}} 
\def\bbR{\mathbb{R}} 
\def\bbP{\mathbb{P}} 
\def\bbQ{\mathbb{Q}} 
\def\bbE{\mathbb{E}}

\def\1n{\mathbf{1}_n}
\def\0{\mathbf{0}}
\def\1{\mathbf{1}}

\def\A{{\bf A}}
\def\B{{\bf B}}
\def\C{{\bf C}}
\def\D{{\bf D}}
\def\E{{\bf E}}
\def\F{{\bf F}}
\def\G{{\bf G}}
\def\H{{\bf H}}
\def\I{{\bf I}}
\def\J{{\bf J}}
\def\K{{\bf K}}
\def\L{{\bf L}}
\def\M{{\bf M}}
\def\N{{\bf N}}
\def\O{{\bf O}}
\def\P{{\bf P}}
\def\Q{{\bf Q}}
\def\R{{\bf R}}
\def\S{{\bf S}}
\def\T{{\bf T}}
\def\U{{\bf U}}
\def\V{{\bf V}}
\def\W{{\bf W}}
\def\X{{\bf X}}
\def\Y{{\bf Y}}
\def\Z{{\bf Z}}

\def\a{{\bf a}}
\def\b{{\bf b}}
\def\c{{\bf c}}
\def\d{{\bf d}}
\def\e{{\bf e}}
\def\f{{\bf f}}
\def\g{{\bf g}}
\def\h{{\bf h}}
\def\i{{\bf i}}
\def\j{{\bf j}}
\def\k{{\bf k}}
\def\l{{\bf l}}
\def\m{{\bf m}}
\def\n{{\bf n}}
\def\o{{\bf o}}
\def\p{{\bf p}}
\def\q{{\bf q}}
\def\r{{\bf r}}
\def\s{{\bf s}}
\def\t{{\bf t}}
\def\u{{\bf u}}
\def\v{{\bf v}}
\def\w{{\bf w}}
\def\x{{\bf x}}
\def\y{{\bf y}}
\def\z{{\bf z}}

\def\balpha{\mbox{\boldmath{$\alpha$}}}
\def\bbeta{\mbox{\boldmath{$\beta$}}}
\def\bdelta{\mbox{\boldmath{$\delta$}}}
\def\bgamma{\mbox{\boldmath{$\gamma$}}}
\def\blambda{\mbox{\boldmath{$\lambda$}}}
\def\bsigma{\mbox{\boldmath{$\sigma$}}}
\def\btheta{\mbox{\boldmath{$\theta$}}}
\def\bomega{\mbox{\boldmath{$\omega$}}}
\def\bxi{\mbox{\boldmath{$\xi$}}}
\def\bnu{\mbox{\boldmath{$\nu$}}}                                  
\def\bphi{\mbox{\boldmath{$\phi$}}}
\def\bmu{\mbox{\boldmath{$\mu$}}}

\def\bDelta{\mbox{\boldmath{$\Delta$}}}
\def\bOmega{\mbox{\boldmath{$\Omega$}}}
\def\bPhi{\mbox{\boldmath{$\Phi$}}}
\def\bLambda{\mbox{\boldmath{$\Lambda$}}}
\def\bSigma{\mbox{\boldmath{$\Sigma$}}}
\def\bGamma{\mbox{\boldmath{$\Gamma$}}}
                                  
\newcommand{\myprob}[1]{\mathop{\mathbb{P}}_{#1}}

\newcommand{\myexp}[1]{\mathop{\mathbb{E}}_{#1}}

\newcommand{\mydelta}[1]{1_{#1}}

\newcommand{\myminimum}[1]{\mathop{\textrm{minimum}}_{#1}}
\newcommand{\mymaximum}[1]{\mathop{\textrm{maximum}}_{#1}}    
\newcommand{\mymin}[1]{\mathop{\textrm{minimize}}_{#1}}
\newcommand{\mymax}[1]{\mathop{\textrm{maximize}}_{#1}}
\newcommand{\mymins}[1]{\mathop{\textrm{min.}}_{#1}}
\newcommand{\mymaxs}[1]{\mathop{\textrm{max.}}_{#1}}  
\newcommand{\myargmin}[1]{\mathop{\textrm{argmin}}_{#1}} 
\newcommand{\myargmax}[1]{\mathop{\textrm{argmax}}_{#1}} 
\newcommand{\myst}{\textrm{s.t. }}

\newcommand{\denselist}{\itemsep -1pt}
\newcommand{\sparselist}{\itemsep 1pt}

\definecolor{pink}{rgb}{0.9,0.5,0.5}
\definecolor{purple}{rgb}{0.5, 0.4, 0.8}   
\definecolor{gray}{rgb}{0.3, 0.3, 0.3}
\definecolor{mygreen}{rgb}{0.2, 0.6, 0.2}

\newcommand{\cyan}[1]{\textcolor{cyan}{#1}}
\newcommand{\red}[1]{\textcolor{red}{#1}}  
\newcommand{\blue}[1]{\textcolor{blue}{#1}}
\newcommand{\magenta}[1]{\textcolor{magenta}{#1}}
\newcommand{\pink}[1]{\textcolor{pink}{#1}}
\newcommand{\green}[1]{\textcolor{green}{#1}} 
\newcommand{\gray}[1]{\textcolor{gray}{#1}}    
\newcommand{\mygreen}[1]{\textcolor{mygreen}{#1}}    
\newcommand{\purple}[1]{\textcolor{purple}{#1}}       

\definecolor{greena}{rgb}{0.4, 0.5, 0.1}
\newcommand{\greena}[1]{\textcolor{greena}{#1}}

\definecolor{bluea}{rgb}{0, 0.4, 0.6}
\newcommand{\bluea}[1]{\textcolor{bluea}{#1}}
\definecolor{reda}{rgb}{0.6, 0.2, 0.1}
\newcommand{\reda}[1]{\textcolor{reda}{#1}}

\def\changemargin#1#2{\list{}{\rightmargin#2\leftmargin#1}\item[]}
\let\endchangemargin=\endlist
                                               
\newcommand{\cm}[1]{}

\newcommand{\mhoai}[1]{{\color{magenta}\textbf{[MH: #1]}}}

\newcommand{\mtodo}[1]{{\color{red}$\blacksquare$\textbf{[TODO: #1]}}}
\newcommand{\myheading}[1]{\vspace{1ex}\noindent \textbf{#1}}
\newcommand{\htimesw}[2]{\mbox{$#1$$\times$$#2$}}



\begin{abstract}


This paper introduces a method to encode the blur operators of an arbitrary dataset of sharp-blur image pairs into a blur kernel space. Assuming the encoded kernel space is close enough to in-the-wild blur operators, we propose an alternating optimization algorithm for blind image deblurring. It approximates an unseen blur operator by a kernel in the encoded space and searches for the corresponding sharp image. Unlike recent deep-learning-based methods, our system can handle unseen blur kernel, while avoiding using complicated handcrafted priors on the blur operator often found in classical methods. Due to the method's design, the encoded kernel space is fully differentiable, thus can be easily adopted in deep neural network models. Moreover, our method can be used for blur synthesis by transferring existing blur operators from a given dataset into a new domain. Finally, we provide experimental results to confirm the effectiveness of the proposed method. The code is available at \url{https://github.com/VinAIResearch/blur-kernel-space-exploring}.

\end{abstract}

\section{Introduction}

Motion blur occurs due to camera shake or rapid movement of objects in a scene. Image deblurring is the task of removing the blur artifacts to improve the quality of the captured image. Image deblurring is an important task with many applications, especially during the current age of mobile devices and handheld cameras. Image deblurring, however, is still an unsolved problem, despite much research effort over the past decades.


Mathematically, the task of image deblurring is to recover the sharp image $x$ given a blurry image $y$. One can assume the below mathematical model that relates $x$ and $y$:
\begin{equation}
    y = \hat{\mathcal{F}}(x, k) + \eta \approx \hat{\mathcal{F}}(x, k),
    \label{eq:generaldeblurring}
\end{equation}
where $\hat{\mathcal{F}}(\cdot, k)$ is the blur operator with the blur kernel $k$, and $\eta$ is noise. In the simplest form, $\hat{\mathcal{F}}(\cdot, k)$ is assumed to be a convolution function with $k$ being a convolution kernel and $\eta$ being white Gaussian noise. Given a blurry image $y$, the deblurring task is to recover the sharp image $x$ and optionally the blur operator $\hat{\mathcal{F}}(\cdot, k)$.



\begin{figure}[t]
    \centering
    \includegraphics[width=0.9\linewidth]{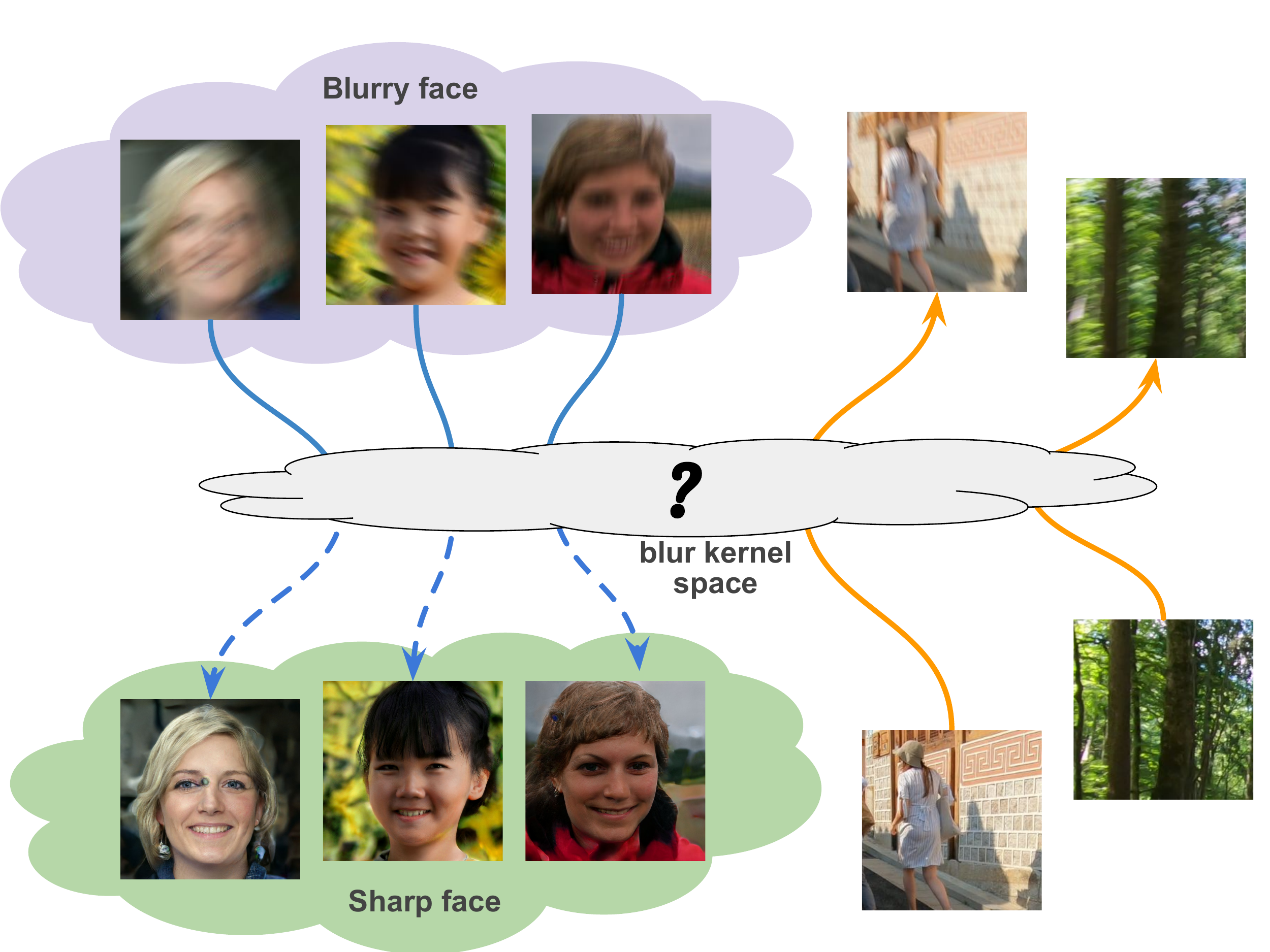}
    \caption{The space of blur kernels is the missing element for successful blur removal and synthesis. Previous image debluring methods either overlooked the importance of this kernel space or made inadequate assumption about it. In this paper, we propose to learn this blur kernel space from a dataset of sharp-blurry image pairs (orange arrows) and leverage this encoded space for image deblurring (blue arrows).}
    \label{fig:generalidea}
\end{figure}

A popular approach to recover the sharp image is to use the  Maximum A Posterior (MAP) estimate. That is to find $x$ and $k$ to maximize the posterior propbability $\bbP(x, k|y)$ assuming $\hat{\mF}$ is known. This is equivalent to optimizing:  
\begin{align}
        x, k = \myargmax{x, k} \bbP(y|x, k) \bbP(x) \bbP(k). \label{eq:map}
\end{align}

However, this is an ill-posed problem and there are infinitely many pairs of $(k, x)$ that lead to the same probability $\bbP(y|x, k)$, so the key aspect of the above MAP approach is to define proper models for the prior distributions $\bbP(x)$ and $\bbP(k)$. In fact, many deblurring methods focus on either designing handcrafted priors for $x$ and $k$ \cite{chan1998total,krishnan2009fast,pan2016blind,liu2014blind} or learning the deep image prior~\cite{ulyanov2018deep, ren2020neural}. However, all of these works assume the blur operator is a convolutional operator, and this assumption does not hold in practice. These MAP-based methods cannot handle complex in-the-wild blur operators and usually produce undesirable artifacts when testing on real-world blurry images.

An alternative approach is to directly learn a function that maps from a blurry image to the corresponding non-blurry image. This function can be a deep convolutional network and the parameters of the network can be learned using paired training data of blurry and non-blurry images~\cite{nah2017deep,tao2018scale,kupyn2018deblurgan,kupyn2019deblurgan}. Unlike the MAP-based approach, this approach learns the inverse function of the blur operator directly without explicitly reasoning about the blur operator and the distribution of the blur kernel. Given the lack of an explicit representation for the space of the blur kernels, this approach does not generalize well beyond the set of individual blur kernels seen during training. This approach~\cite{nah2017deep,tao2018scale,kupyn2018deblurgan,kupyn2019deblurgan} produces poor results when testing on blur operators that are not present in the training set. In our experiments, these deep-learning models degenerate to an identity map when testing on an out-of-domain blur operator; the recovered image is nearly identical to the input image. This is a known issue, and it is referred to as ``the trivial solution'' by traditional deblurring methods.  The MAP-based methods tackle this problem by putting prior distributions on the sharp image and the blur kernel. However, those priors cannot be readily applied to the existing deep-learning models due to the lack of an explicit representation for the blur kernels. 

In this paper, we propose to address the limitations of both aforementioned approaches as follows. First, we devise a deep-learning formulation with an explicit representation for the blur kernel and the blur operator. Second, we use a data-driven approach to learn the family of blur operators and the latent manifold of the blur kernels, instead of assuming that the blur operator is a convolutional operator as used in existing MAP-based methods. Specifically, we simultaneously learn a blur operator family $\mF$ and a blur kernel extractor $\mG$ such that:
\begin{align}
         y = \mathcal{F}(x, k) \quad \textrm{and} \quad
         k = \mathcal{G}(x, y) \label{eq:ourformula}.
\end{align}
Note in this paper, $\mathcal{F}$ is referred to as the blur operator \textsl{family}. For a specific blur kernel $k$,  $\mathcal{F}(\cdot, k)$ is a specific blur operator from the family of blur operators. We call $k$ the blur kernel of the blur operator $\mathcal{F}(\cdot, k)$. When the functional form of $\mF$ is fixed, we will refer to a blur operator $\mathcal{F}(\cdot, k)$ by its blur kernel $k$ if there is no confusion.

Once the blur operator family $\mathcal{F}$ has been learned, we can use it to deblur an input image $y$ by finding $x$ and $k$ to satisfy the above equations using alternating optimization. Moreover, we can incorporate additional constraints on the solution space of $x$ to generate more realistic results. For example, we can use a deep generative model to learn the manifold of natural images and constraint the solution space to this manifold. The conceptual idea is illustrated in \Fref{fig:generalidea}.

Our method can also be used for blur synthesis. This can be done by transferring the blur kernel of a sharp-blurry image pair to another image. Blur synthesis is useful in many ways. For example, we can transfer the real-world motion blur of an existing dataset~\cite{rim_2020_ECCV} to another domain where it might be difficult to collect paired data. Blur synthesis can also be used for training data augmentation, improving the robustness of a downstream task such as face recognition or eye gaze estimation.



In short, the contributions of our paper are: (1) we propose a novel method to encode the blur kernel space for a dataset of blur-sharp image pairs, which can be used to deblur images that contain unseen blur operators; (2) we propose a novel blur synthesis method and demonstrate its utilities; and (3) we obtain state-of-the-art deblurring results on several datasets.



\section{Related Work}
\subsection{Image deblurring}
Image deblurring algorithms can be divided into two main categories: MAP-based and learning-based methods. 

\minisection{MAP-based blind image deblurring.}
In MAP-based methods, finding good priors for the sharp images and blur kernels ($\bbP(x)$ and $\bbP(k)$ in \Eref{eq:map}) are two main focuses. For the sharp images, gradient-based prior is usually adopted since the gradient of natural images is highly sparse. In particular, \citet{chan1998total} proposed a total-variation (TV) penalty that encouraged the sparsity of the image gradient. \citet{krishnan2009fast} suggested that the image gradient followed Hyper-laplacian distribution. However, \citet{levin2009understanding} showed that these gradient-based priors could favor blurry images over sharp ones and lead to the trivial solution, i.e., $x = y$ and $k$ is the identity operator. \citet{krishnan2011blind} used $\ell1/\ell2$ regularization that gave sharp image the lowest penalty. \citet{pan2016blind} showed that the dark channel of a sharp image was usually sparser than the dark channel of the corresponding blurry image. Overall, these priors only model low-level statistics of images, which are neither adequate nor domain-invariant. 

Recently, \citet{ulyanov2018deep} introduced Deep Image Prior (DIP)  for image restoration tasks. A network $G$ was learned so that each image $I$ was represented by a fixed vector $z$ such that $I = G_{\theta}(z)$. \citet{ren2020neural} proposed SelfDeblur method using two DIPs for $x$ and $k$. Instead of using alternating optimization like other MAP-based methods, they jointly sought $x$ and $k$ using a gradient-based optimizer.

All aforementioned methods assumed the blur kernel was linear and uniform, i.e., it can be represented as a convolution kernel. However, this assumption is not true for real-world blur. Non-linear camera response functions can cause non-linear blur kernels while non-uniform blur kernels appear when only a small part of the image moves. There were some attempts for non-uniform deblurring \cite{whyte2012non,cho2007removing,nagy1998restoring,shan2007rotational}, but they still assumed the blur was locally uniform, and they were not very practical given the high computational cost.

\minisection{Learning-based deblurring.}
Many deep deblurring models have been proposed over the past few years. \citet{nah2017deep} proposed a multi-scale network for end-to-end image deblurring. It deblurred an image in three scale levels; the result from the lower level was used as an input of its upper level. Similarly, \citet{tao2018scale} employed a scale-recurrent structure for image deblurring. GAN \cite{goodfellow2014generative} was first used for image deblurring in \cite{kupyn2018deblurgan}, whereas a high-quality image was generated conditioned on the blurry input image. \citet{kupyn2019deblurgan} introduced DeblurGANv2, which used Feature Dynamic Networks \cite{lin2017feature} to extract image features and two discriminators for global and patch levels. DeblurGANv2 achieved impressive run-time while maintaining reasonable results on common benchmarks. There were also works on multi-frame deblurring \cite{wang2019edvr,su2017deep,zhou2019spatio} and domain-specific deblurring \cite{ren2020neural,lin2020learning,Shen_2018_CVPR,song2019joint,xu2017learning,yasarla2020deblurring,hradivs2015convolutional}.

Unfortunately, deep-learning models do not perform well for cross-domain tasks. For example, models trained on the REDS dataset \cite{nah2019ntire} perform poorly on GOPRO \cite{nah2017deep}, despite the visual similarity between the two datasets. As a result,  deep deblurring models have not been used in real-world applications. This kernel overfitting phenomenon has not been explained in prior works.

\subsection{GAN-inversion image restoration}

Image manifolds generated by GANs \cite{goodfellow2014generative} were used to approximate the solution space for image restoration problem in recent works \cite{pan2020exploiting,menon2020pulse}. They sought an image in the manifold such that its degradation version was the closest to the provided low-quality image. The benefits of this method are twofold. First, this method guarantees a sharp and realistic outcome. Meanwhile, image restoration is ill-posed with multiple solutions, and the common image restoration methods often yield a blurry result towards the average of all possible solutions \cite{menon2020pulse}. Second, in the case of blind deblurring, this method bypasses the kernel overfitting issue in deep image restoration models. 

Existing works in this direction, however, just cover simple known degradations such as bicubic downsampling. To handle the challenging in-the-wild motion-blur degradation, we first need to model the family of blur operators.

\subsection{Blur synthesis}
To train deep deblurring models, large-scale and high-quality datasets are needed. But it is hard to capture pairs of corresponding sharp and blurry images in real life, so blur synthesis has been widely used. Assuming uniform blur (i.e., a convolutional blur kernel), a common approach is to synthesize the trajectory of the blur kernel and apply this synthetic kernel on the sharp image set. \citet{chakrabarti2016neural} generated blur trajectories by randomly sampling six points on a grid and connected those points by a spline. \citet{schuler2015learning} sampled blur trajectories by a Gaussian process. These methods could only synthesize uniform blur and they did not take the scene structure into account. Therefore, synthesized blurry images are unrealistic.

More sophisticated blur synthesis algorithms rely on the blur generation process in the camera model. In particular, an image in color space can be modeled as: $I = g \left(\frac{1}{T} \int_0^T S(t)dt\right)$, 
where $S(t)$ is the sensor signal at time $t$, $T$ is the exposure time, and $g$ the camera response function. \citet{nah2017deep} approximated $g$ by the gamma function $g(x) = x^{\frac{1}{\gamma}}$. They converted a frame $I$ to its corresponding signal sensor $g^{-1}(I)$, averaged consucutive frames in that signal domain, then converted it back to the color space. The REDS dataset \cite{nah2019ntire} was synthesized similarly but with an increased video temporal resolution and a more sophisticated camera response function.

To reduce the gap between synthetic and real-world blur, \citet{rim_2020_ECCV} proposed a real-world blur dataset that was captured by two identical cameras with different shutter speeds. However, the data collection process was complicated, requiring elaborate setup with customized hardware.


\section{Methodology}

In this section, we first describe a method to learn the blur operator family $\mF$ that explains the blurs between paired data of sharp-blurry images. We will then explain how the blur operator family can be used for removing or synthesizing blur. 


\subsection{Learning the blur operator family} \label{subsec:encode}

Given a training set of $n$ data pairs $\{(x_i, y_i)\}_{i=1}^{n}$, our goal is to learn a blur operator family that models the blur between the sharp image $x_i$ and the corresponding blurry image $y_i$ for all $i$'s. Each pair is associated with a latent blur kernel $k_i$; and the blurry image $y_i$ is obtained by applying the blur operator family on the sharp image $x_i$ with the blur kernel $k_i$ as parameters, i.e., $y_i = \mF(x_i, k_i)$.  Traditional MAP-based methods often assume  $\mF(\cdot, k_i)$ to be the convolutional operator and $k_i$ a convolutional kernel, but this assumption does not hold for real blurs in the wild.

Learning $\mF$ is challenging because $\{k_i\}$ are latent variables. Fortunately, each $k_i$ is specific to a sharp-blurry image pair, so we can assume $k_i$ can be recovered by a kernel extractor function $\mG$, i.e., $k_i = \mG(x_i, y_i)$. We can  learn both the blur operator family $\mF$ and the kernel extractor $\mG$ by minizing the differences between the synthesized blurry image $\mF(x_i, \mG(x_i, y_i))$ and the actual blurry image $y_i$. In this paper, we implement them by two neural networks, an encoder-decoder with skip connection~\cite{ronneberger2015u} for~$\mathcal{F}$ and a residual network~\cite{he2016deep} for $\mathcal{G}$. Both $\mathcal{F}$ and $\mathcal{G}$ are fully differentiable, and they can be jointly optimized by minimizing the following loss function:
\begin{align}
    \sum_{i=1}^{n} \rho(y_i, \mF(x_i, \mG(x_i, y_i))),
    \label{eq:obj_encode}
\end{align}
where $\rho(\cdot)$ is the Charbonnier loss \cite{lai2017deep} measuring the distance between the ``fake'' blurry image $\mF(x_i, \mG(x_i, y_i))$ and the corresponding real blurry image $y_i$.

\begin{figure}[t]
    \centering
    \includegraphics[width=0.9\linewidth]{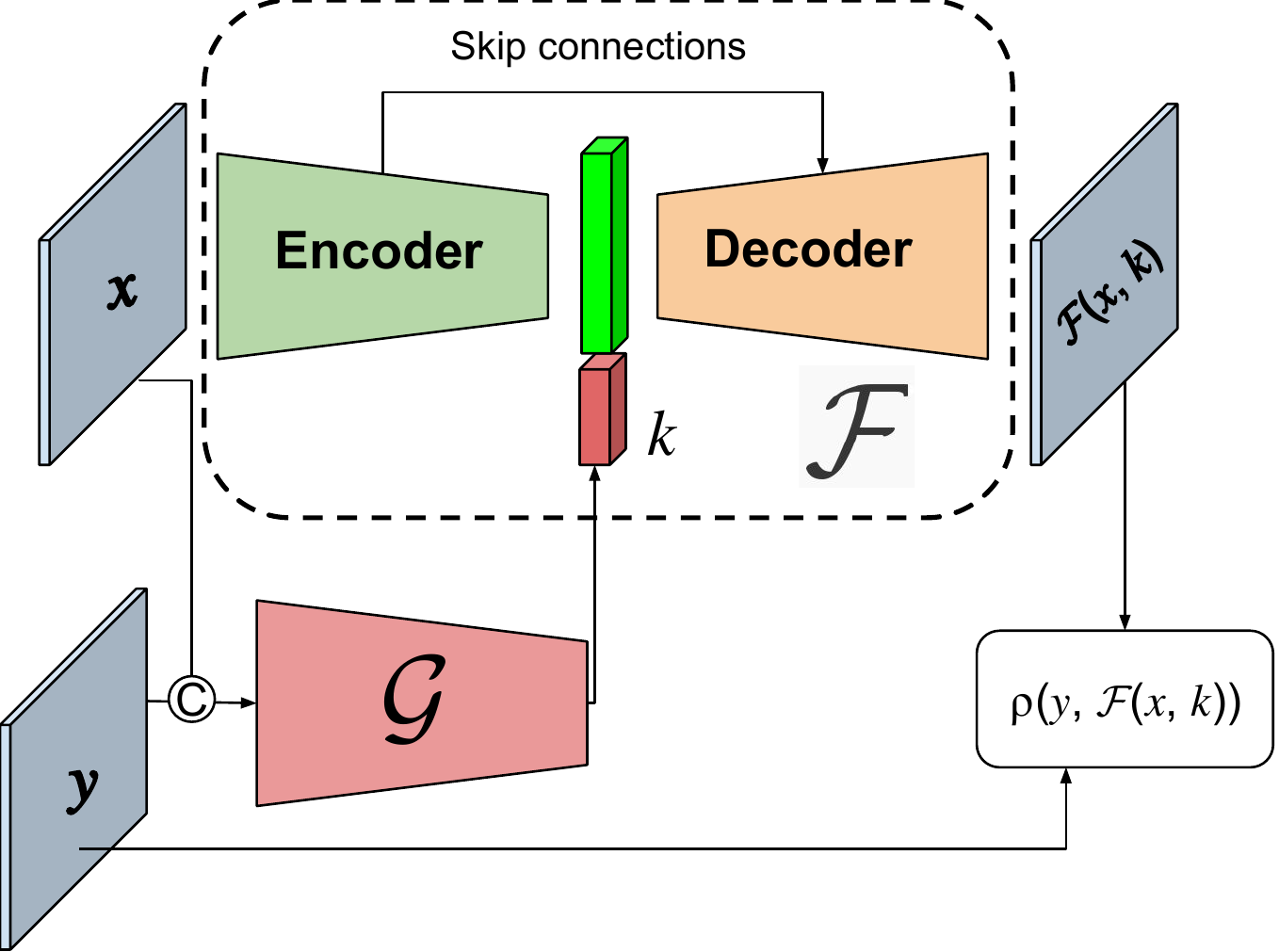}
    \caption{{\bf Roles of the blur operator family $\mF$ and the blur kernel extractor $\mG$ and their architectures}. $\mG$ can be used to extract the blur kernel $k$, while $\mF$ can be used to generate a blurry image given the blur kernel $k$. $\mF$ is an encoder-decoder network with skip connection, while $\mG$ is a residual network. } \label{fig:general_architecture}
\end{figure}

This procedure is illustrated in \Fref{fig:general_architecture}. First, we sample $(x, y)$ from a dataset of image pairs. Second, we fit the concatenation of these images into $\mathcal{G}$ to generate the corresponding encoded blur kernel vector $k$. Third, with $x$ and $k$ as the input, we use $\mathcal{F}$ to create the synthesized blurry image. $\mF$ encodes $x$ into a bottle-neck embedding vector, concatenates that embedding vector with $k$, and decodes it to get the synthesized blurry image.
Details of the architecture choices and hyper-parameters tuning are given in the supplementary materials.

\begin{algorithm}[t]
    \caption{Blind image deblurring}
    \label {algo:imagedeblurring}
    
    \textbf{Input:} blurry image $y$ \\
    \textbf{Output:} sharp image $x$
    \begin{algorithmic}[1]
        \State Sample $z_x \sim \mathcal{N}(0, I)$
        \State Randomly initialize $\theta_x$ of $G^{x}_{\theta_x}$
    
        \While{$\theta_x$ has not converged}
            \State Sample $z_k \sim \mathcal{N}(0, I)$    
            \State Randomly initialize $\theta_k$ of $G^{k}_{\theta_k}$
            \While{$\theta_k$ has not converged}
                \State $g_k \leftarrow \partial \mathcal{L}(\theta_x, \theta_k) / \partial \theta_k$
                \State $\theta_k \leftarrow \theta_k + \alpha * ADAM(\theta_k, g_k)$
            \EndWhile

            \State $g_x \leftarrow \partial \mathcal{L}(\theta_x, \theta_k) / \partial \theta_x$
            \State $\theta_x \leftarrow \theta_x + \alpha * ADAM(\theta_x, g_x)$
        \EndWhile
        
        \State $x = G_{\theta_x}(z_x)$
    \end{algorithmic}
\end{algorithm}

        

    
        
    
        

\subsection{Blind image deblurring}
\label{subsec:imagedeblurring}


Once the blur operator family $\mF$ has been learned, we can use it for image deblurring. Given a blurry image $y$, our task is to recover the sharp image $x$. We pose it as the optimization problem, where we seek to recover both the sharp image $x$ and the blur kernel $k$ to minimize  $\rho(y, \mF(x, k))$. To optimize $\rho(y, \mF(x, k))$, we propose an iterative optimization procedure that alternates between the following two steps: (A) fix the blur kernel $k$ and optimize the latent sharp image $x$, and (B) fix $x$ and optimize for $k$. 

To stablize the optimization process and to obtain better deblurring results, we propose to add a couple of regularization terms into the objective function and reparameterize both $x$ and $k$ with Deep Image Prior (DIP) \cite{ulyanov2018deep} as follows. First, we propose to add a regularization term on the $L_2$ norm of the kernel $k$ to stablize the optimization process and avoid the trivial solution. Second, we propose to 
use the Hyper-Laplacian prior \cite{krishnan2009fast} on the image gradients of $\x$ to encourage the sparsity of the gradients, reducing noise and creating more natural looking image $x$. This corresponds to adding the regularization term: $(g_u^2(x) + g_v^2(x))^{\alpha/2}$ into the objective function, where $g_u$ and $g_v$ are the horizontal and vertical derivative operators respectively. Adding the regularization terms leads to the updated objective: 
\begin{align}
    \rho(y, \mathcal{F}(x, k)) + \lambda||k||_2 + \gamma (g_u^2(x) + g_v^2(x))^{\alpha/2}, 
\end{align}
where $\lambda, \gamma, \alpha$ are tunable hyper-parameters. 

Finally, inspired by the success of Deep Image Prior \cite{ulyanov2018deep} for zero-shot image restoration \cite{ulyanov2018deep,ren2020neural,gandelsman2019double,liu2019image}, we propose to reparameterize both $x$ and $k$ by neural networks. In particular, instead of optimizing $x$ directly, we take $x$
as the stochastic output of a neural network $G_{\theta_x}^x$ and we optimize the parameters $\theta_x$ of the network instead. Specifically, we define $x = G^{x}_{\theta_x}(z_x)$, where $z_x$ is standard normal random vector, i.e., $z_x \sim \mN(0, I)$. Similarly, we reparameterize $k = G^{k}_{\theta_k}(z_k)$. The final objective function for deblurring is: 
\begin{align}
    \mL(\theta_x, \theta_k)&=\rho(y, \mathcal{F}(x, k)) + \lambda||k||_2 + \gamma (g_u^2(x) + g_v^2(x))^{\alpha/2} \nonumber \\
    \textrm{where }   x &= G^{x}_{\theta_x}(z_x), z_x \sim \mN(0,I), \\
      k &= G^{k}_{\theta_k}(z_k), z_k \sim \mN(0,I).
\end{align}
This objective function can be optimized using \Aref{algo:imagedeblurring}.

\subsection{Approximated manifold of natural images}
\label{subsec:restrictssolutionspace}


In \Sref{subsec:imagedeblurring}, we propose a general solution for image deblurring, where little assumption is made about the space of the sharp image $x$. We use DIP to reparameterize $x$ as the output of a neural network with schotastic input, and we optimize the parameter of the network instead. However, in many situations, the domain of the sharp image $x$ is simpler, e.g., being a face or a car. In this situation, we can have better reparameterization for $x$, taking into account the learned manifold for the specific domain of $x$. 


In this paper, we also consider the image manifold proposed by \citet{menon2020pulse}. We reparameterize $x$ by $G_{style}(z)$ in which $G_{style}$ is the pretrained StyleGAN \cite{karras2019style}, $z$ is optimized along the sphere $\sqrt{d}S^{d-1}$ using spherical projected gradient descent~\cite{menon2020pulse}.

\begin{figure*}[t]
    \setlength{\tabcolsep}{0.3pt}
    \small
    \begin{center}
    \begin{tabular}{cccccc}
        \multicolumn{1}{c}{Blur} & 
        \multicolumn{1}{c}{SelfDeblur \cite{ren2020neural}} & 
        \multicolumn{1}{c}{DeblurGANv2 \cite{kupyn2019deblurgan}} & 
        \multicolumn{1}{c}{SRN-Deblur \cite{tao2018scale}} & 
        \multicolumn{1}{c}{Ours} &
        \multicolumn{1}{c}{Sharp}\\
        \cellimgsixperrow{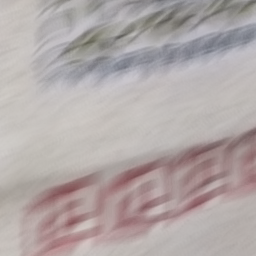} &
        \cellimgsixperrow{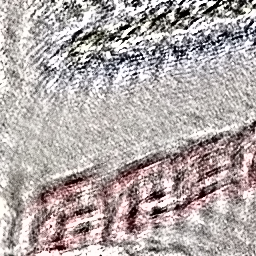} &
        \cellimgsixperrow{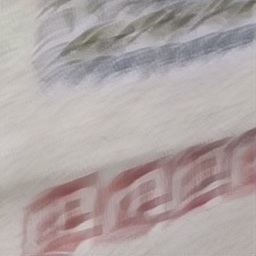} &
        \cellimgsixperrow{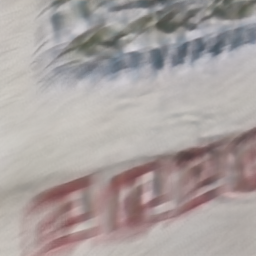} &
        \cellimgsixperrow{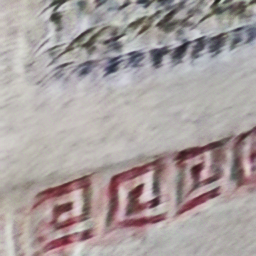} &
        \cellimgsixperrow{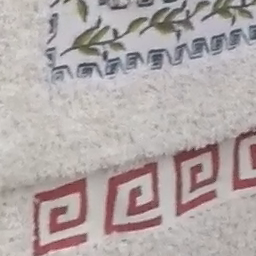}\\
        0.489 & 0.630 & 0.442 & 0.448 & 0.348 &\\
        \cellimgsixperrow{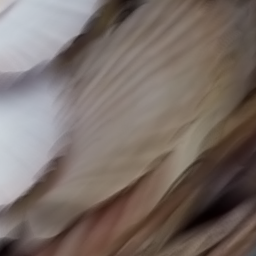} &
        \cellimgsixperrow{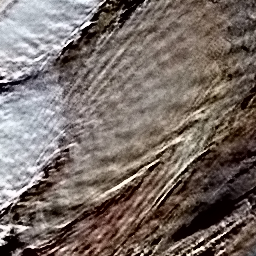} &
        \cellimgsixperrow{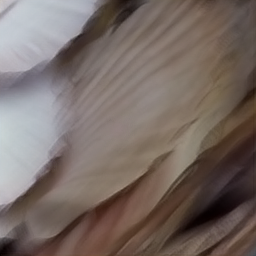} &
        \cellimgsixperrow{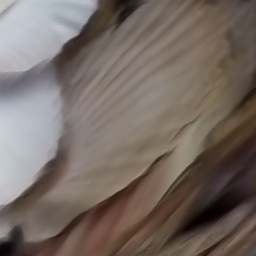} &
        \cellimgsixperrow{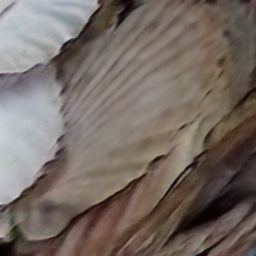} &
        \cellimgsixperrow{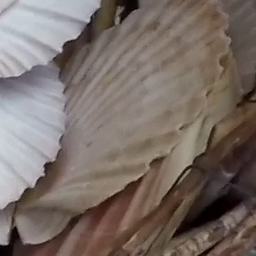}\\
        0.630 & 0.857 & 0.663 & 0.633 & 0.601\\
        \cellimgsixperrow{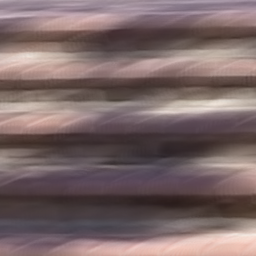} &
        \cellimgsixperrow{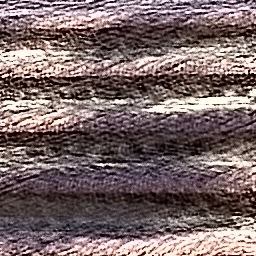} &
        \cellimgsixperrow{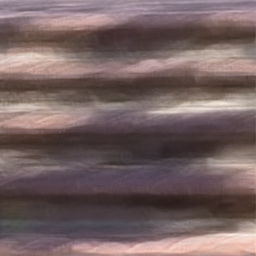} &
        \cellimgsixperrow{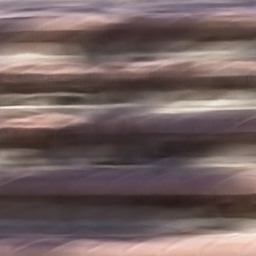} &
        \cellimgsixperrow{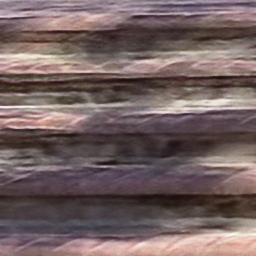} &
        \cellimgsixperrow{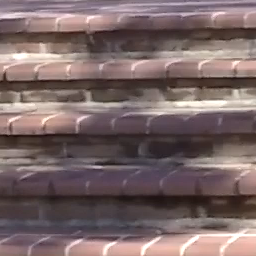}\\
        0.717 & 0.780 & 0.707 & 0.694 & 0.664\\
    \end{tabular}
    \end{center}
    \vskip -0.15in
    \caption{Results of deblurring methods trained on REDS and tested on GOPRO, and their LPIPS score \cite{zhang2018unreasonable} (lower is better).} 
    \label{fig:imagedeblurring}
\end{figure*}

\begin{figure*}[t]
    \setlength{\tabcolsep}{0.3pt}
    \small
    \begin{center}
    \begin{tabular}{lllllll}
        \multicolumn{1}{c}{Blur} & 
        \multicolumn{1}{c}{SelfDeblur \cite{ren2020neural}} & 
        \multicolumn{1}{c}{\cite{kupyn2019deblurgan} REDS} & 
        \multicolumn{1}{c}{\cite{kupyn2019deblurgan} imgaug} & 
        \multicolumn{1}{c}{\cite{tao2018scale} REDS} & 
        \multicolumn{1}{c}{\cite{tao2018scale} imgaug} & 
        \multicolumn{1}{c}{Ours}\\
        \cellimgsmall{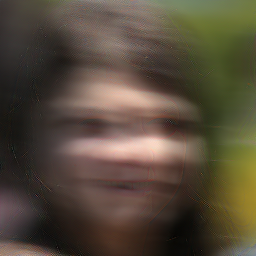} &
        \cellimgsmall{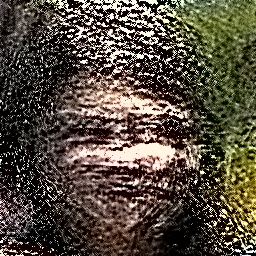} &
        \cellimgsmall{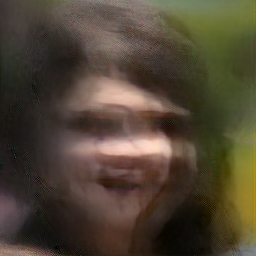} &
        \cellimgsmall{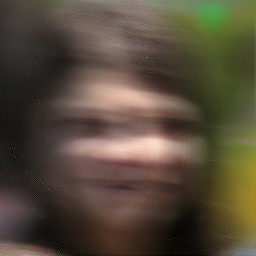} &
        \cellimgsmall{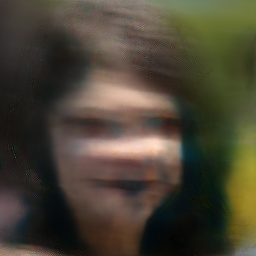} &
        \cellimgsmall{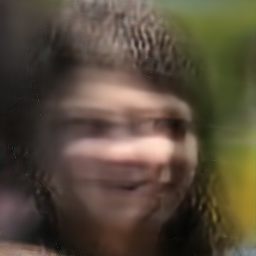} &
        \cellimgsmall{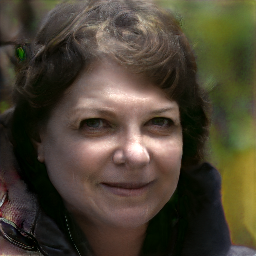}\\
        \cellimgsmall{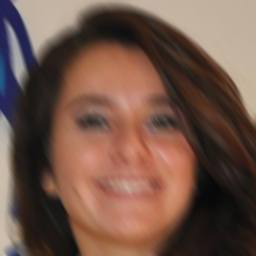} &
        \cellimgsmall{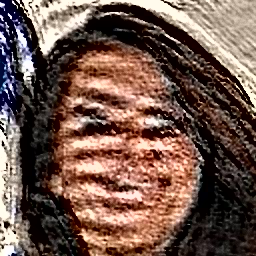} &
        \cellimgsmall{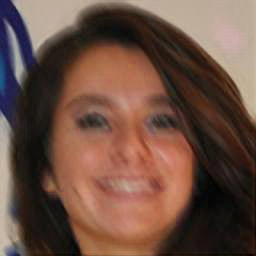} &
        \cellimgsmall{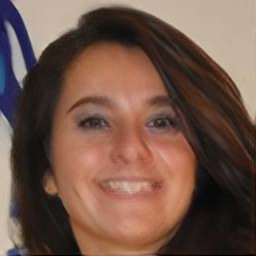} &
        \cellimgsmall{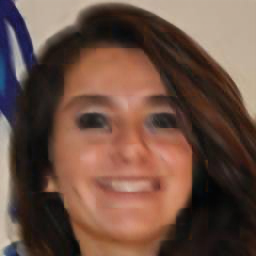} &
        \cellimgsmall{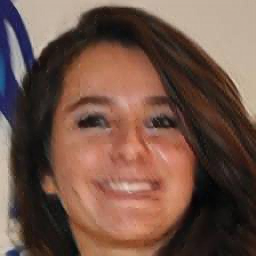} &
        \cellimgsmall{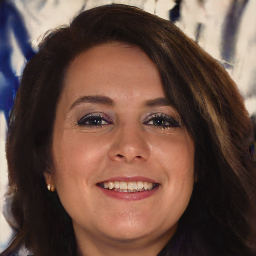}\\
        \cellimgsmall{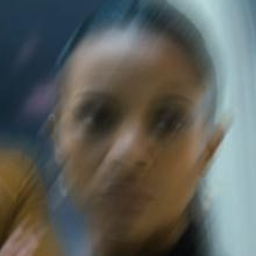} &
        \cellimgsmall{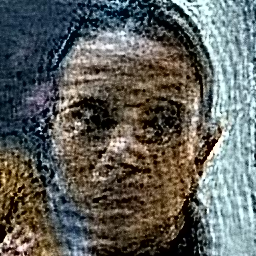} &
        \cellimgsmall{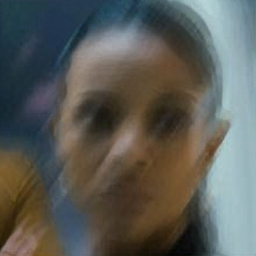} &
        \cellimgsmall{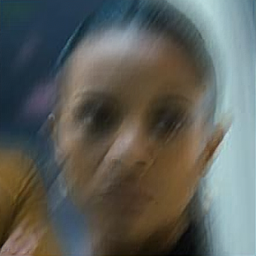} &
        \cellimgsmall{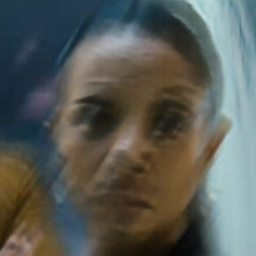} &
        \cellimgsmall{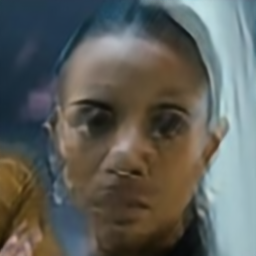} &
        \cellimgsmall{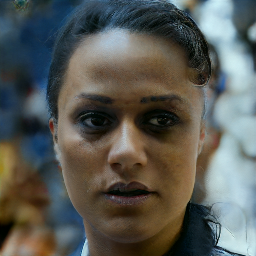}\\
        \cellimgsmall{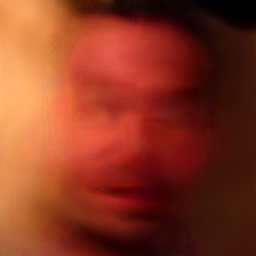} &
        \cellimgsmall{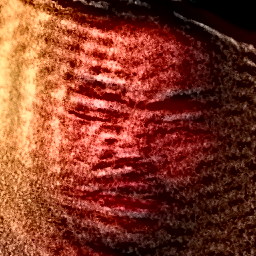} &
        \cellimgsmall{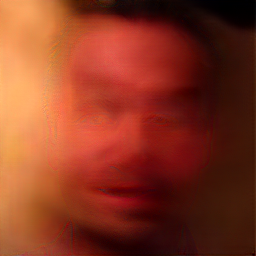} &
        \cellimgsmall{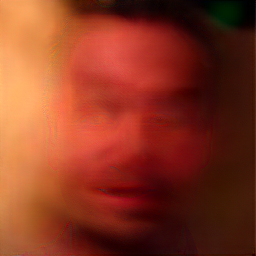} &
        \cellimgsmall{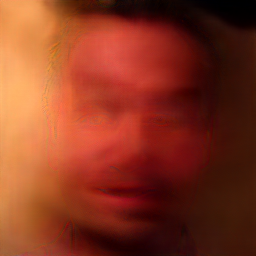} &
        \cellimgsmall{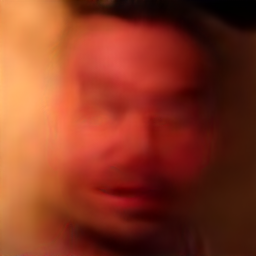} &
        \cellimgsmall{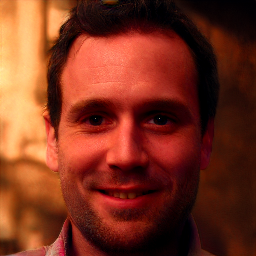}\\
    \end{tabular}
    \end{center}
    \vskip -0.15in
    \caption{Qualitative results of deblurring methods. Here DeblurGANv2 REDS is the model trained with face dataset using REDS kernel, while DeblurGANv2 imgaug is the model trained with face dataset using imgaug. The blurry image in the first and second rows are synthesized using blur transferring technique in \Sref{fig:augmentation} and \textit{imgaug} \cite{imgaug} respectively. The last two rows are in-the-wild blurry images that we randomly collect on the Internet.}
    \label{fig:face}
    \vspace{-2mm}
\end{figure*}

\subsection{Blur synthesis using blur transferring \label{subsec:augment}}

There exist datasets of paired images with ``close-to-real'' blurs, such as REDS \cite{nah2019ntire}, GOPRO \cite{nah2017deep}, or real-world blur \cite{rim_2020_ECCV}. But the collection of these datasets required elaborate setups, expensive hardware (e.g., high-speed camera), and enormous effort. Unfortunately, similar datasets do not exist for many application domains (e.g., faces and scene text), and it is difficult or even impossible to replicate these laboratory setups to collect data for in-the-wild environments (e.g., street scenes). 

To this end, a benefit of our approach is the ability to transfer the motion blurs from an existing dataset to a new set of images. In particular, given a dataset with pairs of sharp-blurry images, we can first train $\mathcal{F}$ and $\mathcal{G}$ as described in \Sref{subsec:encode}. To transfer the motion blur between the image pair $(x, y)$ to a new image $\hat{x}$, we can simply compute: $\hat{y} := \mF(\hat{x}, \mG(x, y))$.

\section{Experiments}
We perform extensive experiments to verify the effectiveness of our blur kernel encoding method. We also provide results for image deblurring and blur synthesis. All the experiments are conducted on a single NVidia V100 GPU. Image deblurring experiments are cross-domain. In particular, all data-driven methods are trained on the REDS dataset~\cite{nah2019ntire} and tested on the GOPRO dataset \cite{nah2017deep}.


\myheading{REDS dataset \cite{nah2019ntire}} comprises 300 high-quality videos with various scenes. The videos are captured at 120fps. The corresponding blurry videos are synthesized by upsampling the frame rate and averaging the neighboring frames. We use this dataset to train our kernel extractor as well as deep deblurring models.

\myheading{GOPRO dataset \cite{nah2017deep}} consists 3142 sharp-blur pair of frames. Those frames are captured at 240fps. The synthesis process is similar to REDS dataset, except for the choice of the camera response function. We use this dataset to test the deblurring methods.

\myheading{Levin dataset \cite{levin2009understanding}} is generated using eight convolution kernels with different sizes. Here we use its kernels to synthesize uniform blur on other datasets.

\myheading{FFHQ dataset \cite{karras2019style}} is a human face dataset. This dataset consists of 70,000 high-quality $1024{\times}1024$ images with various genders, ethics, background, and accessories. This dataset was used to train the StyleGAN model.

\myheading{CelebA-HQ dataset} \cite{karras2017progressive} is a human face dataset that consists of 30,000 images at $1024{\times}1024$ resolution. Its images were selected from the CelebA dataset \cite{liu2015faceattributes}, but the quality was improved using some preprocessing steps such as JPEG removal and $4{\times}$ super-resolution.

\subsection{Blur kernel extractor}
This section verifies if our blur kernel extractor can accurately extract and transfer blur from a sharp-blurry image pair to another image. We use the known explicit kernels from the Levin dataset to synthesize blurry images in training and testing for experiments with ground-truth labels. As for experiments on datasets without explicit blur kernels, such as REDs and GOPRO, we check the stability of the deblurring networks trained on internal blur-swapped data.

\subsubsection{Testing blur kernel encoding on Levin dataset}

Suppose we have a ground-truth blur operator family $\hat{\mathcal{F}}$. We train $\mathcal{F}$ and $\mathcal{G}$ using a sharp-blur pair dataset generated by $\hat{\mathcal{F}}$. Then we can measure the performance of the blur kernel extractor by calculating the distance between $\mathcal{F}(x, \mathcal{G}(x, y))$ and $\hat{\mathcal{F}}(x, h)$ for arbitrary pair $(x, h)$ and $y = \hat{\mathcal{F}}(x, h)$. 

In this experiment, we let $\hat{\mathcal{F}}(\cdot, h)$ be a convolutional operator whose kernel is one of the eight used in the Levin dataset \cite{levin2009understanding}. To generate training data, we randomly select 5000 sharp images from the REDS dataset \cite{nah2019ntire} and generate 5000 corresponding blurry images using the mentioned kernels. Then we use these 5000 pairs to learn $\mathcal{F}$ and $\mathcal{G}$. To create testing data, we randomly sample two other disjointed image sets $S$ and $T$ for the source and target sharp images in blur transfer. Each set consists of 500 sharp images from GOPRO dataset \cite{nah2017deep}. Then for each testing kernel $k$, we generate the blur images in the source set $y_k = \hat{\mathcal{F}}(x, k) = k * x$, apply blur from $(x, y_k)$ to each $\hat{x} \in T$ via the trained $\mathcal{F}$ and $\mathcal{G}$, and compute the average PSNR score.
    \begin{align}
        \frac{\sum_{x \in S, \hat{x} \in T} PSNR (\mathcal{F}(\hat{x}, \mathcal{G}(x, y_k)), \hat{\mathcal{F}}(\hat{x}, k))}{|S|\times|T|}.
    \end{align}


\begin{table}[hb]
    \centering
    \vspace{-2mm}
    \begin{tabular}{ccccc}
        \toprule
         & kernel 1 & kernel 2 & kernel 3 & kernel 4\\
         \cmidrule(lr){2-5}
         PSNR (db) & 49.48 & 51.93 & 52.06 & 53.74\\
         \bottomrule
         & kernel 5 & kernel 6 & kernel 7 & kernel 8 \\
         \cmidrule(lr){2-5}
         PSNR (db) & 49.91 & 49.49 & 51.43 & 50.38\\
         \bottomrule
    \end{tabular}
    \vskip 0.05in
    \caption{Results of our blur kernel extraction on Levin dataset}
    \label{tab:levinexp}
\end{table}

We report the test results in  \Tref{tab:levinexp}. Our method achieves very high PSNR scores, demonstrating its ability to extract and transfer the blur kernels. 

\subsubsection{Training on synthetic datasets}
For a sharp-blur dataset without explicit blur kernels, we can randomly swap the blur operator between its pairs using our method. To be more specific, for each sharp-blur pair $(x, y)$ and a random sharp image $\hat{x}$ from this dataset, we generate the blurry image $\hat{y}$ using the blur kernel extracted from $(x, y)$. Then we use this synthetic dataset to train a deep deblurring model and compare its performance to the one trained on the original dataset. In this experiment, we choose SRN-Deblur \cite{tao2018scale}, a typical deep image deblurring method. The testing datasets are REDS and GOPRO. 

The performance of deblurring networks, measured by the average PSNR score on test sets, is reported in \Tref{tab:synexp}. PSNR scores when training on blur-swapped datasets are comparable to the ones obtained when training on the original dataset.


\setlength{\tabcolsep}{20pt}
\begin{table}[ht]
    \centering
    \begin{tabular}{lcc}
        \toprule
          & \multicolumn{2}{c}{Dataset}\\
         \cmidrule(lr){2-3}
         Training data & REDS & GOPRO \\
         \midrule
         Original & 30.70 & 30.20 \\
         Blur-swapped & 29.43& 28.49 \\
         \bottomrule
    \end{tabular}
    \vskip 0.05in
    \caption{Results of SRN-Deblur trained \cite{tao2018scale} on the original and blur-swapped datasets.}
    \label{tab:synexp}
\end{table}

\subsection{General blind image deblurring}
\subsubsection{Qualitative results}
We now evaluate our blind image deblurring method, described in \Sref{subsec:imagedeblurring}, and compare it to other methods in a cross domain setting. We use the state-of-the-art deep-learning-based methods, including DeblurGANv2~\cite{kupyn2019deblurgan}, SRN-Deblur \cite{tao2018scale}, and a recent kernel-based algorithm called SelfDeblur \cite{ren2020neural}. We train all the methods using REDS dataset \cite{nah2019ntire} and test them on GOPRO dataset \cite{nah2017deep}.

Some visualization results and their corresponding LPIPS scores \cite{zhang2018unreasonable} are shown in \Fref{fig:imagedeblurring}. The methods based on deep neural networks  \cite{kupyn2019deblurgan,tao2018scale} produce results that are very similar to the input. On the other hand, the predicted images of  SelfDeblur \cite{ren2020neural} are noisy with many artifacts. Our method consistently generates sharp and visually pleasing results.

\subsubsection{Retrieving unseen kernel}\label{sec:exp_retrieve}
Our algorithm is based on the assumption that an unseen blur operator can be well approximated using the encoded blur kernel space. Here we conduct an experiment to verify this assumption. We use $\mathcal{F}$ and $\mathcal{G}$ that are trained on one dataset, either REDS or GOPRO, to retrieve unseen blur operator of each sharp-blur image pair in the testing subset of the same or different dataset using step (B) in \Sref{subsec:imagedeblurring}. To evaluate the accuracy of that extracted blur, we compute PSNR score between the reconstructed and original blurry images. The average PSNR score for each configuration is reported in  \Tref{tab:retrieve}. As can be seen, the quality of kernels extracted in cross-domain setting is similar to the ones in same-domain configuration. It shows that our method is effective in handling unseen blur.

\begin{figure}[ht]
    \small
    \begin{center}
    \begin{tabular}{ccc}
        sharp & original blur & retrieved blur\\
        \cellimgsmall{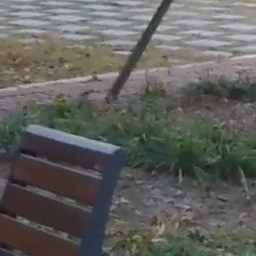} &
        \cellimgsmall{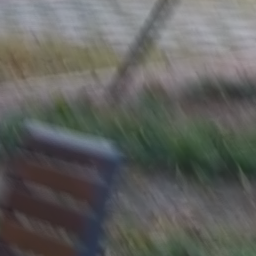} &
        \cellimgsmall{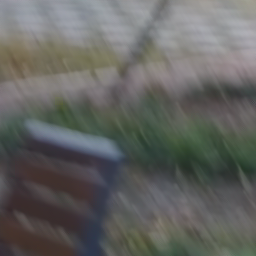}\\
        \cellimgsmall{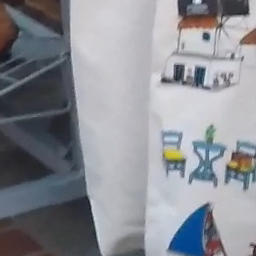} &
        \cellimgsmall{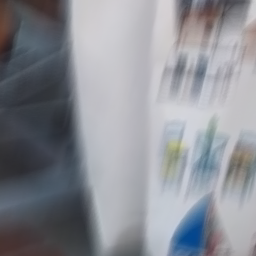} &
        \cellimgsmall{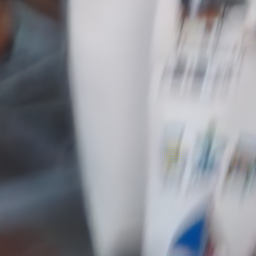}\\
    \end{tabular}
    \end{center}
    \vskip -0.1in
    \caption{Retrieving unseen kernel. The first column shows the sharp images from the GOPRO dataset, the second column shows their corresponding blurry images. In the last row, we approximate the blur operators using the kernels from REDS dataset and apply it to the sharp images.}
    \label{fig:retrieving}
\end{figure}

\Fref{fig:retrieving} visualizes some results when training on REDS and testing on GOPRO. Our reconstructed blurry images are close to the original ones, indicating the high quality of the extracted kernels.

\setlength{\tabcolsep}{3pt}

\setlength{\tabcolsep}{20pt}
\begin{table}[ht]
    \centering
    \begin{tabular}{lcc}
        \toprule
          & \multicolumn{2}{c}{Test set}\\
         \cmidrule(lr){2-3}
         Tranining set & REDS4 & GOPRO \\
         \midrule
         REDS & 34.35 & 30.67\\
         GOPRO & 31.38 & 35.13\\
         \bottomrule
    \end{tabular}
    \vskip 0.05in
    \caption{Results of our method in retrieving unseen blur kernel with same and cross-domain configs.}
    \label{tab:retrieve}
\end{table}

\subsection{Using an approximated natural image manifold}
\subsubsection{Qualitative results}
As discussed in \Sref{subsec:restrictssolutionspace}, we can incorporate a GAN-based image manifold as the sharp image prior to attain realistic deblurring results. Following \cite{menon2020pulse}, we conduct face deblurring experiments using the StyleGAN model pretrained on the FFHQ dataset to approximate the natural facial image manifold. We use both synthesized and in-the-wild blurry images for testing. As for synthetic data, we use images from CelebHQ dataset \cite{karras2017progressive}. The blur synthesis techniques include motion-blur augmentation from the \textit{imgaug} (the second row in \Fref{fig:face}) tool \cite{imgaug} and the blur transferred from the GOPRO dataset (the first row in \Fref{fig:face}). As for in-the-wild images, we search for blurry faces from the Internet (the last two rows in \Fref{fig:face}). Each deep model is trained using FFHQ dataset \cite{karras2019style} with blur operators are synthesized by \textit{imgaug} or blur kernels transferred from GOPRO dataset \cite{nah2017deep}. As for our method, we use the blur extractor trained on REDS dataset in \Sref{sec:exp_retrieve}. All the test blurs, therefore, are unseen to our method.

We compare our deblurring results and different baseline methods in \Fref{fig:face}. As can be seen, the deep deblurring models \cite{kupyn2019deblurgan,tao2018scale} fail to produce sharp outcomes, particularly on unseen blur. The state-of-the-art MAP-based algorithm \cite{ren2020neural} generates unrealistic and noisy images. In contrast, our method can successfully approximate realistic sharp face outputs in all test cases.

\subsubsection{Loss convergence}
\begin{figure}[ht]
    \begin{center}
        \includegraphics[scale=0.5]{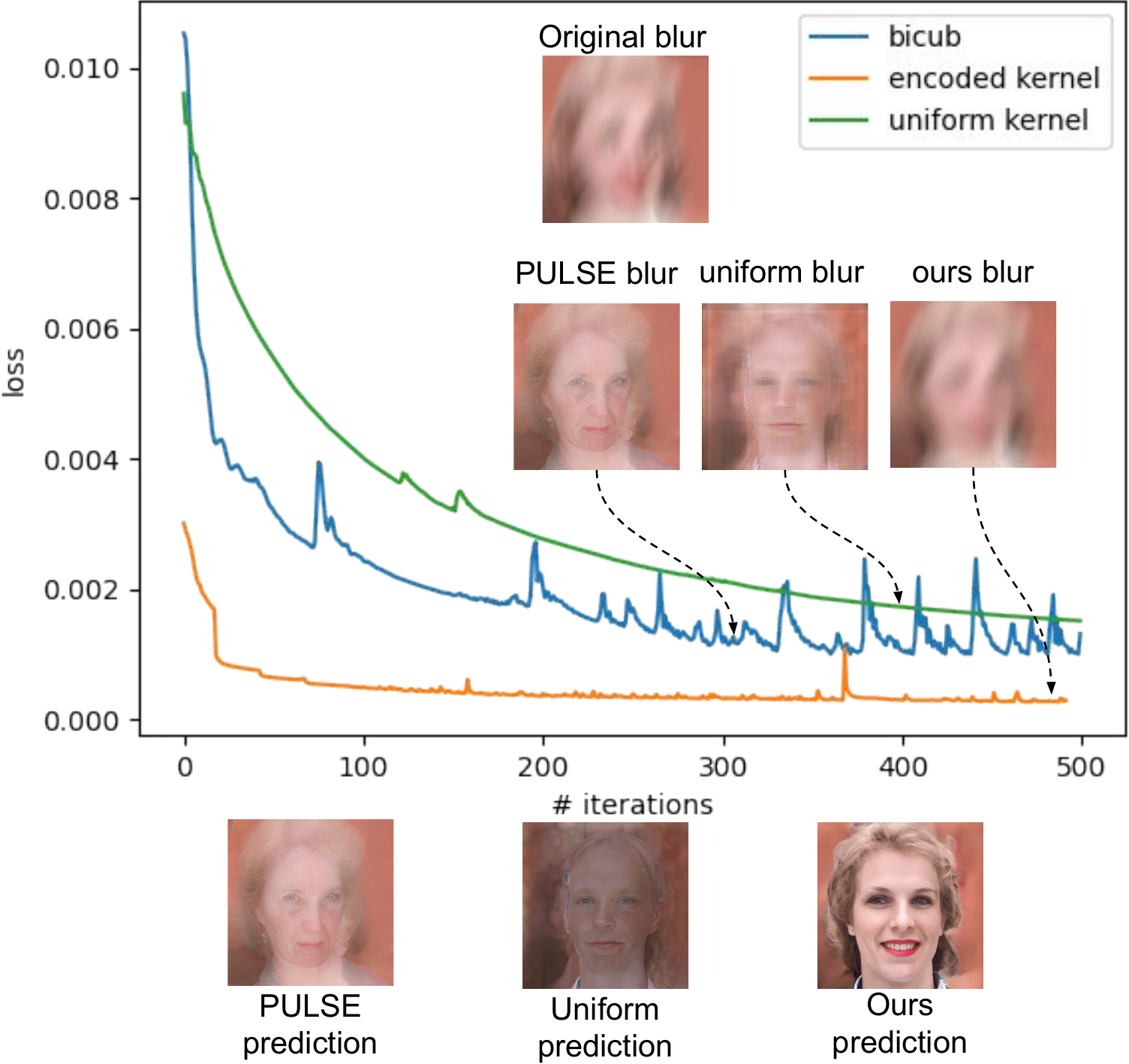}
        \caption{Loss convergence of the method in \Sref{subsec:restrictssolutionspace} when using different kernel priors.}
        \label{fig:loss_convergence}
    \end{center}
    \vspace{-5mm}
\end{figure}

One may think that the good deblurring results in the previous experiment are purely due to restricting the sharp image solution space to a GAN manifold. Yes, but the blur kernel prior is equally important; without a good blur kernel prior, the method would fail to converge to desirable results. To prove it, we analyze the optimization processes on a specific deblurring example with different blur kernel manifolds: (1) the traditional convolution kernels with DIP used in SelfDeblur \cite{ren2020neural}, (2) the bicubic downsampling kernel used in PULSE \cite{menon2020pulse}, and (3) our encoded kernel. The results are shown in \Fref{fig:loss_convergence}. The first two methods failed to converge since the real blur operator is neither linear nor uniform. In contrast, the method using our kernel method quickly converges to a realistic face.

\subsection{Blur synthesis}
\begin{figure}
    \begin{center}
        \begin{tabular}{cccc}
            $\hat{x}$ & $x$ & $y$ & $\hat{y}$\\
            \cellimgtiny{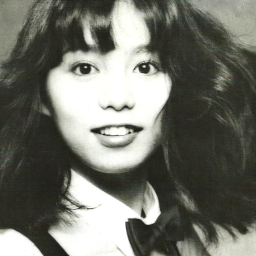} &
            \cellimgtiny{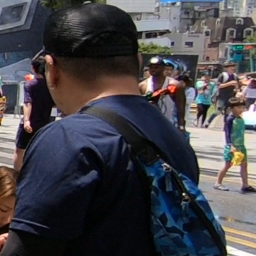} &
            \cellimgtiny{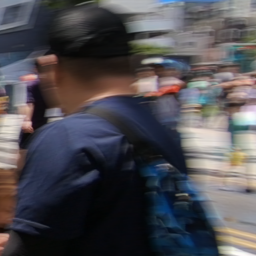} &
            \cellimgtiny{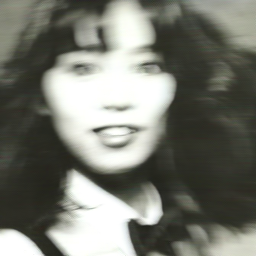}\\
            \cellimgtiny{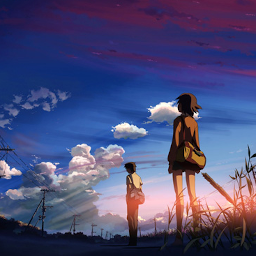} &
            \cellimgtiny{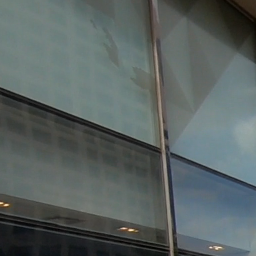} &
            \cellimgtiny{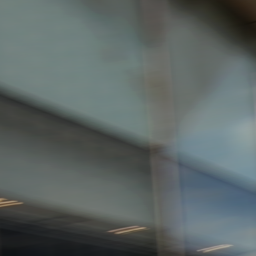} &
            \cellimgtiny{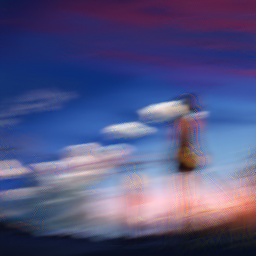}\\
        \end{tabular}
    \caption{Transfering blur kernel from the source pair $x, y$ to the target sharp $\hat{x}$ to generate the target blurry image $\hat{y}$.}
    \label{fig:augmentation}
    \end{center}
    \vspace{-5mm}
\end{figure}

Our blur transfer method is effective in synthesizing new blurry images. In \Fref{fig:augmentation}, we transfer the blur operator from the source sharp-blur pair $(x, y)$ (the two middle columns) to the target sharp image $\hat{x}$ (the first column) to synthesize its corresponding blurry image $\hat{y}$. We see that the content of $\hat{x}$ is fully preserved in $\hat{y}$, and the blur in $\hat{y}$ looks similar to the blur in $y$. Our method can also work with any type of images, such as grayscale images (the first row) or animation images (the second row). 

One application of this blur synthesis is data augmentation. We experiment with the use of this augmentation technique to improve image deblurring. In particular, we use FFHQ dataset \cite{karras2019style} to synthesize three sharp-blur datasets with different types of blur kernels: (1) common motion-blur kernels generated by imgaug tool \cite{imgaug}, (2) our encoded REDS kernels, and (3) our encoded GOPRO kernels. The first dataset is the traditional deblurring dataset. The second dataset can be considered as data augmentation, and the last dataset is used for unseen blur testing. We train SRN-Deblur models \cite{tao2018scale} in two scenarios: using only the first dataset or using the combination of the first two datasets. Testing results are reported in \Tref{tab:blursynFFHQ}. The network trained on the combined data is more stable and performs better in the unseen blur scenario.





\setlength{\tabcolsep}{7pt}
\begin{table}[ht]
    \centering
    \begin{tabular}{lccc}
        \toprule
          & \multicolumn{3}{c}{Test kernels}\\
          \cmidrule(lr){2-4}
         Tranining kernels & imgaug & REDS & GOPRO\\
         \midrule
         imgaug & \textbf{28.64} & 24.22 & 22.96\\
         comb. & 28.30 & \textbf{28.37} & \textbf{23.92}\\
         \bottomrule
    \end{tabular}
    \vskip 0.05in
    \caption{Effect of blur augmentation in improving SRN-Deblur \cite{tao2018scale} model, tested on the synthetic FFHQ datasets.}
    \label{tab:blursynFFHQ}
    \vspace{-4mm}
\end{table}



\setlength{\tabcolsep}{0pt}

\section{Conclusion}
In this paper, we have proposed a method to encode the blur kernel space of an arbitrary dataset of sharp-blur image pairs and leverage this encoded space to solve some specific tasks such as image deblurring and blur synthesis. For image deblurring, we have shown that our method can handle unseen blur operators. For blur synthesis, our method can transfer blurs from a given dataset of sharp-blur image pairs into any domain of interest, including domains of facial, grayscale, and animated images.

{\small
\setlength{\bibsep}{0pt}

}

\end{document}


\title{Explore Image Deblurring via Encoded Blur Kernel Space\\
--- Supplementary material ---}

\author{
Phong Tran$^{1}$ \quad Anh Tuan Tran$^{1,2}$ \quad Quynh Phung$^{1}$ \quad Minh Hoai$^{1,3}$ \\
$^1$VinAI Research, Hanoi, Vietnam,
$^2$VinUniversity, Hanoi, Vietnam,\\
$^3$Stony Brook University, Stony Brook, NY 11790, USA\\
{\tt\small \{v.phongtt15,v.anhtt152,v.quynhpt29,v.hoainm\}@vinai.io}
}

\maketitle

\def\mA{\mathcal{A}}
\def\mB{\mathcal{B}}
\def\mC{\mathcal{C}}
\def\mD{\mathcal{D}}
\def\mE{\mathcal{E}}
\def\mF{\mathcal{F}}
\def\mG{\mathcal{G}}
\def\mH{\mathcal{H}}
\def\mI{\mathcal{I}}
\def\mJ{\mathcal{J}}
\def\mK{\mathcal{K}}
\def\mL{\mathcal{L}}
\def\mM{\mathcal{M}}
\def\mN{\mathcal{N}}
\def\mO{\mathcal{O}}
\def\mP{\mathcal{P}}
\def\mQ{\mathcal{Q}}
\def\mR{\mathcal{R}}
\def\mS{\mathcal{S}}
\def\mT{\mathcal{T}}
\def\mU{\mathcal{U}}
\def\mV{\mathcal{V}}
\def\mW{\mathcal{W}}
\def\mX{\mathcal{X}}
\def\mY{\mathcal{Y}}
\def\mZ{\mathcal{Z}} 

\def\bbN{\mathbb{N}} 
\def\bbR{\mathbb{R}} 
\def\bbP{\mathbb{P}} 
\def\bbQ{\mathbb{Q}} 
\def\bbE{\mathbb{E}}

\def\1n{\mathbf{1}_n}
\def\0{\mathbf{0}}
\def\1{\mathbf{1}}

\def\A{{\bf A}}
\def\B{{\bf B}}
\def\C{{\bf C}}
\def\D{{\bf D}}
\def\E{{\bf E}}
\def\F{{\bf F}}
\def\G{{\bf G}}
\def\H{{\bf H}}
\def\I{{\bf I}}
\def\J{{\bf J}}
\def\K{{\bf K}}
\def\L{{\bf L}}
\def\M{{\bf M}}
\def\N{{\bf N}}
\def\O{{\bf O}}
\def\P{{\bf P}}
\def\Q{{\bf Q}}
\def\R{{\bf R}}
\def\S{{\bf S}}
\def\T{{\bf T}}
\def\U{{\bf U}}
\def\V{{\bf V}}
\def\W{{\bf W}}
\def\X{{\bf X}}
\def\Y{{\bf Y}}
\def\Z{{\bf Z}}

\def\a{{\bf a}}
\def\b{{\bf b}}
\def\c{{\bf c}}
\def\d{{\bf d}}
\def\e{{\bf e}}
\def\f{{\bf f}}
\def\g{{\bf g}}
\def\h{{\bf h}}
\def\i{{\bf i}}
\def\j{{\bf j}}
\def\k{{\bf k}}
\def\l{{\bf l}}
\def\m{{\bf m}}
\def\n{{\bf n}}
\def\o{{\bf o}}
\def\p{{\bf p}}
\def\q{{\bf q}}
\def\r{{\bf r}}
\def\s{{\bf s}}
\def\t{{\bf t}}
\def\u{{\bf u}}
\def\v{{\bf v}}
\def\w{{\bf w}}
\def\x{{\bf x}}
\def\y{{\bf y}}
\def\z{{\bf z}}

\def\balpha{\mbox{\boldmath{$\alpha$}}}
\def\bbeta{\mbox{\boldmath{$\beta$}}}
\def\bdelta{\mbox{\boldmath{$\delta$}}}
\def\bgamma{\mbox{\boldmath{$\gamma$}}}
\def\blambda{\mbox{\boldmath{$\lambda$}}}
\def\bsigma{\mbox{\boldmath{$\sigma$}}}
\def\btheta{\mbox{\boldmath{$\theta$}}}
\def\bomega{\mbox{\boldmath{$\omega$}}}
\def\bxi{\mbox{\boldmath{$\xi$}}}
\def\bnu{\mbox{\boldmath{$\nu$}}}                                  
\def\bphi{\mbox{\boldmath{$\phi$}}}
\def\bmu{\mbox{\boldmath{$\mu$}}}

\def\bDelta{\mbox{\boldmath{$\Delta$}}}
\def\bOmega{\mbox{\boldmath{$\Omega$}}}
\def\bPhi{\mbox{\boldmath{$\Phi$}}}
\def\bLambda{\mbox{\boldmath{$\Lambda$}}}
\def\bSigma{\mbox{\boldmath{$\Sigma$}}}
\def\bGamma{\mbox{\boldmath{$\Gamma$}}}
                                  
\newcommand{\myprob}[1]{\mathop{\mathbb{P}}_{#1}}

\newcommand{\myexp}[1]{\mathop{\mathbb{E}}_{#1}}

\newcommand{\mydelta}[1]{1_{#1}}

\newcommand{\myminimum}[1]{\mathop{\textrm{minimum}}_{#1}}
\newcommand{\mymaximum}[1]{\mathop{\textrm{maximum}}_{#1}}    
\newcommand{\mymin}[1]{\mathop{\textrm{minimize}}_{#1}}
\newcommand{\mymax}[1]{\mathop{\textrm{maximize}}_{#1}}
\newcommand{\mymins}[1]{\mathop{\textrm{min.}}_{#1}}
\newcommand{\mymaxs}[1]{\mathop{\textrm{max.}}_{#1}}  
\newcommand{\myargmin}[1]{\mathop{\textrm{argmin}}_{#1}} 
\newcommand{\myargmax}[1]{\mathop{\textrm{argmax}}_{#1}} 
\newcommand{\myst}{\textrm{s.t. }}

\newcommand{\denselist}{\itemsep -1pt}
\newcommand{\sparselist}{\itemsep 1pt}

\definecolor{pink}{rgb}{0.9,0.5,0.5}
\definecolor{purple}{rgb}{0.5, 0.4, 0.8}   
\definecolor{gray}{rgb}{0.3, 0.3, 0.3}
\definecolor{mygreen}{rgb}{0.2, 0.6, 0.2}

\newcommand{\cyan}[1]{\textcolor{cyan}{#1}}
\newcommand{\red}[1]{\textcolor{red}{#1}}  
\newcommand{\blue}[1]{\textcolor{blue}{#1}}
\newcommand{\magenta}[1]{\textcolor{magenta}{#1}}
\newcommand{\pink}[1]{\textcolor{pink}{#1}}
\newcommand{\green}[1]{\textcolor{green}{#1}} 
\newcommand{\gray}[1]{\textcolor{gray}{#1}}    
\newcommand{\mygreen}[1]{\textcolor{mygreen}{#1}}    
\newcommand{\purple}[1]{\textcolor{purple}{#1}}       

\definecolor{greena}{rgb}{0.4, 0.5, 0.1}
\newcommand{\greena}[1]{\textcolor{greena}{#1}}

\definecolor{bluea}{rgb}{0, 0.4, 0.6}
\newcommand{\bluea}[1]{\textcolor{bluea}{#1}}
\definecolor{reda}{rgb}{0.6, 0.2, 0.1}
\newcommand{\reda}[1]{\textcolor{reda}{#1}}

\def\changemargin#1#2{\list{}{\rightmargin#2\leftmargin#1}\item[]}
\let\endchangemargin=\endlist
                                               
\newcommand{\cm}[1]{}

\newcommand{\mhoai}[1]{{\color{magenta}\textbf{[MH: #1]}}}

\newcommand{\mtodo}[1]{{\color{red}$\blacksquare$\textbf{[TODO: #1]}}}
\newcommand{\myheading}[1]{\vspace{1ex}\noindent \textbf{#1}}
\newcommand{\htimesw}[2]{\mbox{$#1$$\times$$#2$}}



\begin{abstract}
In this supplement material, we further provide some material including architecture details, hyper-parameter tuning, and qualitative results to further analyze our method.
\end{abstract}


\section{Training environment}
We implement the proposed method using Pytorch 1.4.0. The experiments are conducted using a single Nvidia RTX 2080 Ti GPU. We train $\mathcal{F}$ and $\mathcal{G}$ for $6 \times 10^6$ iterations on either REDS \cite{nah2019ntire} or GOPRO \cite{nah2017deep} datasets. 

\section{Architecture choices}
\begin{figure*}[t]
    \centering
    \includegraphics[scale=0.39]{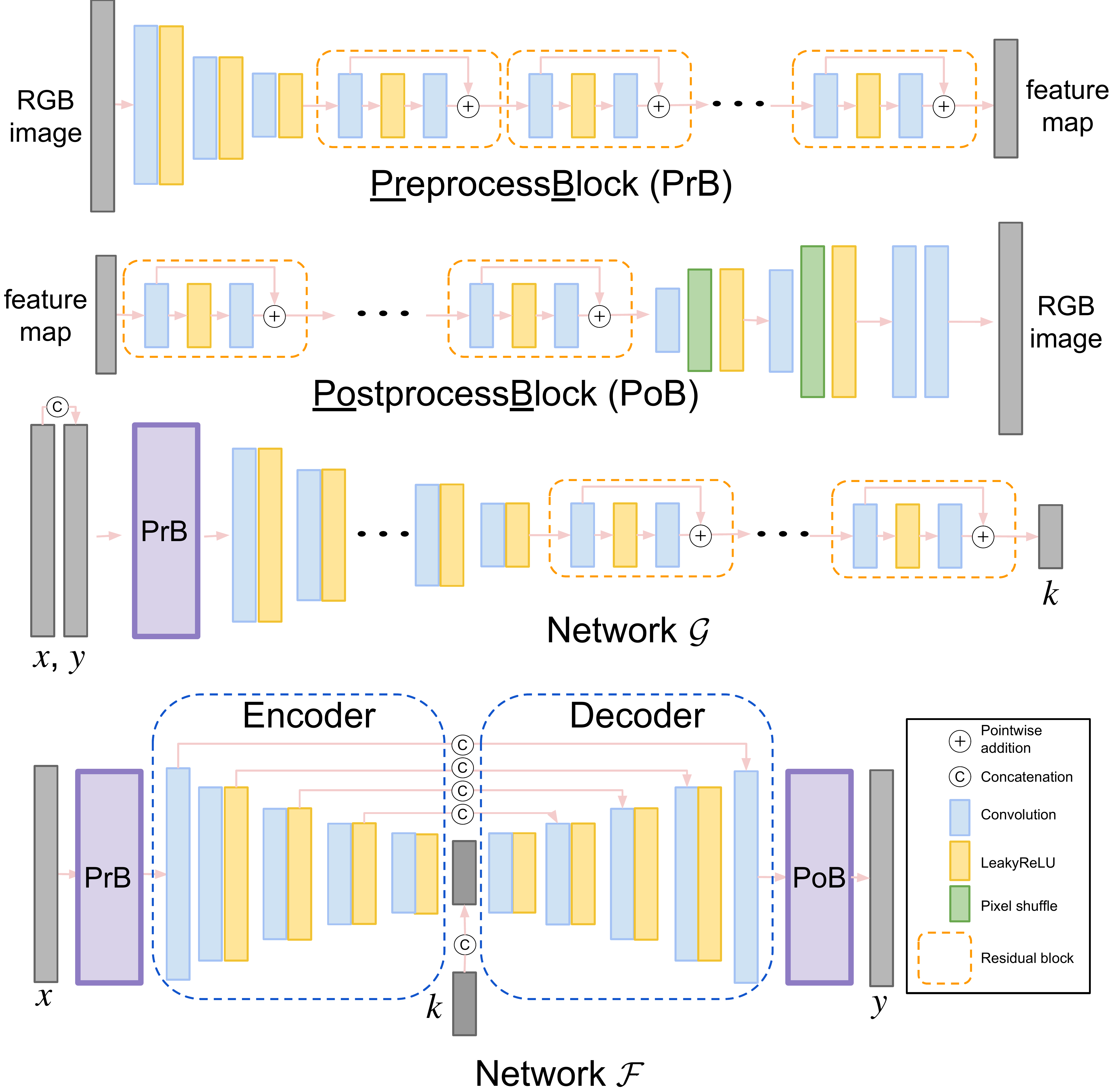}
    \caption{\large Detailed architecture of the proposed method}
    \label{fig:detail_architectures}
\end{figure*}

Here we illustrate in detail the architecture of each component in our proposed method. Details of the overall network are given in \Fref{fig:detail_architectures}

\subsection{Preprocessing and Postprocessing blocks}
We follow a common practice by using a pre-processing block that downsamples the input image twice and convert it to a feature map of size $64 \times H/4 \times W/4$ at the beginning of both $\mathcal{F}$ and $\mathcal{G}$. We denote it as \underline{Pr}eprocess\underline{B}lock or \textbf{PrB} in short. At the end of $\mathcal{F}$, we apply a post-processing block that converts the output feature map of size $64 \times H/4 \times W/4$ back to the image domain. This block is called \underline{Po}stprocess\underline{B}lock, and denoted as \textbf{PoB}.
Their architectures are illustrated in \Tref{tab:kernel_extractor} and \Tref{tab:reconstructor} respectively.

\begin{table}[ht]
    \setlength{\tabcolsep}{6pt}
    \centering
    \begin{tabular}{ll}
        \toprule
        Layer                                    &  Output shape\\
        \midrule
        $\text{Conv}(3, 64, 3, 1, 1)$            &  $64 \times H \hphantom{/1} \times W$\\
        $\text{Conv}(64, 64, 3, 2, 1)$           &  $64 \times H/2 \times W/2$\\
        $\text{Conv}(64, 64, 3, 2, 1)$           &  $64 \times H/4 \times W/4$\\
        $\text{ResBlock}(64) \times 10$          &  $64 \times H/4 \times W/4$\\
        \bottomrule
    \end{tabular}
    \caption{Structure of PreprocessBlock.}
    \label{tab:kernel_extractor}
\end{table}

\begin{table}[ht]
    \setlength{\tabcolsep}{6pt}
    \centering
    \begin{tabular}{ll}
        \toprule
        Layer                                       &  Output shape\\
        \midrule
        $\text{ResBlock}(64) \times 20$             &  $64\hphantom{0} \times H/4 \times W/4$\\
        $\text{Conv}(64, 256, 3, 1, 1)$             &  $64\hphantom{0} \times H/4 \times W/4$\\
        $\text{PixelShuffle}(2)$                    &  $64\hphantom{0} \times H/2 \times W/2$\\
        $\text{LeakyReLU}(0.1)$                     &  $64\hphantom{0} \times H/2 \times W/2$\\
        $\text{Conv}(64, 256, 3, 1, 1)$             &  
        $256 \times H/2 \times W/2$\\
        $\text{PixelShuffle}(2)$                    &  $64\hphantom{0} \times H\hphantom{/1} \times W$\\
        $\text{LeakyReLU}(0.1)$                     &  $64\hphantom{0} \times H\hphantom{/1} \times W$\\
        $\text{Conv}(64, 64, 3, 1, 1)$              &  $64\hphantom{0} \times H\hphantom{/1} \times W$\\
        $\text{Conv}(64, 3, 3, 1, 1)$               &  
        $3\hphantom{00}\times H\hphantom{/1} \times W$\\
        
        \bottomrule
    \end{tabular}
    \caption{Structure of PostprocessBlock.}
    \label{tab:reconstructor}
\end{table}

\begin{table}[ht]
    \setlength{\tabcolsep}{6pt}
    \centering
    \begin{tabular}{ll}
        \toprule
        Layer                                       &  Output shape\\
        \midrule
        PreprocessBlock & $64\hphantom{0} \times H/4\hphantom{00} \times W/4\hphantom{00}$\\
        $\text{Conv}(128, 64, 7, 1, 1)$              &  $64\hphantom{0} \times H/4\hphantom{00} \times W/4$\\
        $\text{LeakyReLU(0.1)}$                      & $64\hphantom{0} \times H/4\hphantom{00} \times W/4$\\
        
        $\text{Conv}(64, 128, 3, 2, 1)$              & 
        $128 \times H/8\hphantom{00} \times W/8$\\
        $\text{LeakyReLU(0.1)}$                      & 
        $128 \times H/8\hphantom{00} \times W/8$\\
        
        $\text{Conv}(128, 256, 3, 2, 1)$             & 
        $256 \times H/16\hphantom{0} \times W/16$\\
        $\text{LeakyReLU(0.1)}$                      & 
        $256 \times H/16\hphantom{0} \times W/16$\\
        
        $\text{Conv}(256, 512, 3, 2, 1)$             & 
        $512 \times H/32\hphantom{0} \times W/32$\\
        $\text{LeakyReLU(0.1)}$                      & 
        $512 \times H/32\hphantom{0} \times W/32$\\
        
        $\text{Conv}(512, 512, 3, 2, 1)$             & 
        $512 \times H/64\hphantom{0} \times W/64$\\
        $\text{LeakyReLU(0.1)}$                      & 
        $512 \times H/64\hphantom{0} \times W/64$\\
        
        $\text{Conv}(512, 512, 3, 2, 1)$             & 
        $512 \times H/128 \times W/128$\\
        $\text{LeakyReLU(0.1)}$                      & 
        $512 \times H/128 \times W/128$\\
        
        $\text{ResBlock}(512) \times 4$              & 
        $512 \times H/128 \times W/128$\\
        
        \bottomrule
    \end{tabular}
    \caption{Structure of $\mathcal{G}$}
    \label{tab:G_architecture}
\end{table}

\begin{table}[ht]
    \setlength{\tabcolsep}{3pt}
    \centering
    \begin{tabular}{l|l}
        \toprule
        \multicolumn{2}{c}{Encoder}\\
        \midrule
        Layer                                        &  Output shape\\
        \midrule
        PreprocessBlock & $64\hphantom{0} \times H/4\hphantom{00} \times W/4\hphantom{00}$\\
        $\text{Conv}(64, 64, 3, 2, 1)$               &  
        $64\hphantom{0} \times H/8\hphantom{00} \times W/8$\\
        $\text{LeakyReLU}(0.1)$                      &  
        $64\hphantom{0} \times H/8\hphantom{00} \times W/8$\\
        $\text{Conv}(64, 128, 3, 2, 1)$              &  
        $128 \times H/16\hphantom{0} \times W/16$\\
        $\text{LeakyReLU}(0.1)$                      &  
        $128 \times H/16\hphantom{0} \times W/16$\\
        $\text{Conv}(128, 256, 3, 2, 1)$             &  
        $256 \times H/32\hphantom{0} \times W/32$\\
        $\text{LeakyReLU}(0.1)$                      &  
        $256 \times H/32\hphantom{0} \times W/32$\\
        $\text{Conv}(256, 512, 3, 2, 1)$             &  
        $512 \times H/64\hphantom{0} \times W/64$\\
        $\text{LeakyReLU}(0.1)$                      &  
        $512 \times H/64\hphantom{0} \times W/64$\\
        $\text{Conv}(512, 512, 3, 2, 1)$             &  
        $512 \times H/128 \times W/128$\\
        $\text{LeakyReLU}(0.1)$                      &  
        $512 \times H/128 \times W/128$\\
        \midrule
        \multicolumn{2}{c}{Decoder}\\
        \midrule
        Layer                                        &  Output shape\\
        \midrule
        $\text{TransConv}(1024, 512, 3, 2, 1)$       &  
        $512 \times H/64 \times W/64$\\
        $\text{LeakyReLU}(0.1)$                      &  
        $512 \times H/64 \times W/64$\\
        $\text{TransConv}(1024, 256, 3, 2, 1)$       &  
        $256 \times H/32 \times W/32$\\
        $\text{LeakyReLU}(0.1)$                      &  
        $256 \times H/32 \times W/32$\\
        $\text{TransConv}(512, 128, 3, 2, 1)$        &  
        $128 \times H/16 \times W/16$\\
        $\text{LeakyReLU}(0.1)$                      &  
        $128 \times H/16 \times W/16$\\
        $\text{TransConv}(256, 64, 3, 2, 1)$         &  
        $64\hphantom{0} \times H/8\hphantom{0} \times W/8$\\
        $\text{LeakyReLU}(0.1)$                      &  
        $64\hphantom{0} \times H/8\hphantom{0} \times W/8$\\
        $\text{TransConv}(128, 64, 3, 2, 1)$         &  
        $64\hphantom{0} \times H/4\hphantom{0} \times W/4$\\
        $\text{LeakyReLU}(0.1)$                      &  
        $64\hphantom{0} \times H/4\hphantom{0} \times W/4$\\
        PostprocessBlock & $64\hphantom{0} \times H\hphantom{000} \times W\hphantom{00}$\\
        
        \bottomrule
    \end{tabular}
    \caption{Structure of the encoder and decoder of $\mathcal{F}$}
    \label{tab:F_architecture}
\end{table}

\subsection{Architecture of $\mathcal{G}$}
We use $\mathcal{G}$ to extract the blur kernel $k$ from a given sharp-blur pair of images $x, y$. We implement $\mathcal{G}$ using the mentioned PreprocessBlock and a follow-up residual neural network \cite{he2016deep}. Input of $\mathcal{G}$ is the concatenation of $x$ and $y$. Its output is a blur kernel of size $512 \times H/128 \times W/128$. Details of its architecture are given in \Tref{tab:G_architecture}.

\subsection{Architecture of $\mathcal{F}$}
$\mathcal{F}$ takes two inputs, the sharp image $x$ and the blur kernel from $\mathcal{G}(x, y)$. As mentioned, $\mathcal{F}$ uses a PreprocessBlock at the beginning and a PostprocessBlock at the end. Between these blocks, we use an encoder-decoder with skip connection \cite{ronneberger2015u}. The encoder downsamples the pre-processed feature map five times and flattens to an embedding vector. This vector is then concatenated with $k$ and fed into a decoder that reconstructs the output feature map. Details of its architecture are illustrated in \Tref{tab:F_architecture}.

\subsection{Architectures of Deep Image Prior}
We adopt the architecture of $\mathcal{G}$ for the network of DIP of the blur kernel. The input $z_k$ is a normal-distributed random tensor with the size equal to the size of the input of $\mathcal{G}$.

For DIP for image, we adopt a U-net \cite{ronneberger2015u} as suggested in \cite{ulyanov2018deep}. The input $z_x$ is a normal-distributed random tensor with size $1 \times 64 \times 64$.

\section{Hyper-parameters tuning}
We trained the networks with an Adam optimizer \cite{kingma2014adam} with $\beta_1$ and $\beta_2$ are $0.9$ and $0.99$ respectively. The initial learning rate was $10^{-4}$ with cosine annealing scheduler \cite{loshchilov2016sgdr} was applied. We set the weight of kernel regularization $||k||_2$ to $6 \times 10^{-4}$ for all image debluring experiments. The weight of Hyper-Laplacian prior \cite{krishnan2009fast} was set to $2 \times 10^{-2}$.

\section{Cross-dataset experiment}
Here we provide quantitative comparisons on GOPRO and HIDE dataset \cite{HAdeblur} in \Tref{tab:hideexp}. We train the model using GOPRO dataset and test on HIDE dataset and vice versa. To make the testing sets, we randomly sample 500 images from each GOPRO and HIDE testing set. Qualitative results are given in \Fref{fig:nonuniform}.

\setlength{\tabcolsep}{4pt}
\begin{table}[ht]
    \centering
    \begin{tabular}{cccc}
        \toprule
          & DeblurGANv2 [15] & SRN-Deblur [36] & ours\\
          \midrule
          GOPRO & 24.35 & 25.21 & \textbf{26.17}\\
          HIDE & 24.65 & 25.25 & \textbf{25.97}\\
         \bottomrule
    \end{tabular}
    \vskip 0.05in
    \caption{PSNR scores of deblurring methods on the HIDE and GOPRO datasets.}
    \vspace{-5mm}
    \label{tab:hideexp}
\end{table}

\begin{figure}[ht]
    \setlength{\tabcolsep}{0.3pt}
    \small
    \begin{center}
    \begin{tabular}{cccc}
        \cellimg{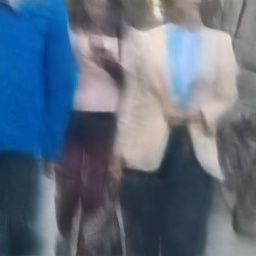} &
        \cellimg{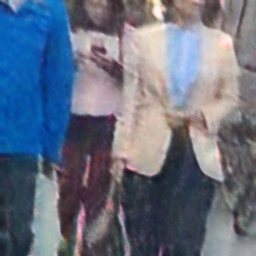} & 
    \end{tabular}
    \end{center}
    \vskip -0.15in
    \caption{Deblurring results (left: SRN, right: ours) on HIDE dataset}
    \label{fig:nonuniform}
\end{figure}

\section{Inference time}
We trained the kernel extractor using an Nvidia V100 with 5GB memory. It took $600K$ iterations to converge (about 4 days). The average inference time for a $256{\times}256$ image using an Nvidia V100 is 209.53s.

\section{More qualitative results}
\begin{figure*}
    \huge
    \begin{center}
        \begin{tabular}{cccc}
            $\hat{x}$ & $x$ & $y$ & $\hat{y}$\\
            \cellimg{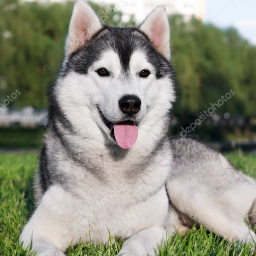} &
            \cellimg{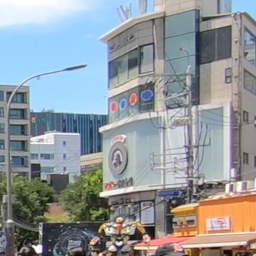} &
            \cellimg{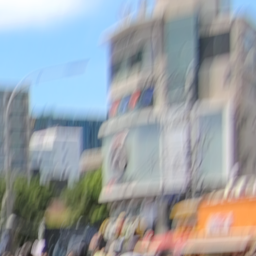} &
            \cellimg{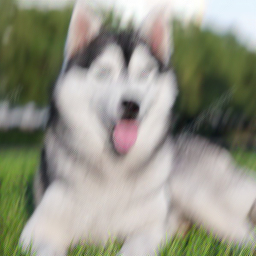}\\
            \cellimg{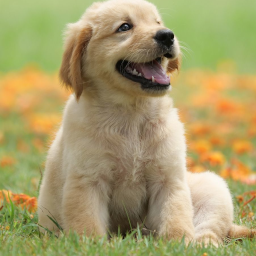} &
            \cellimg{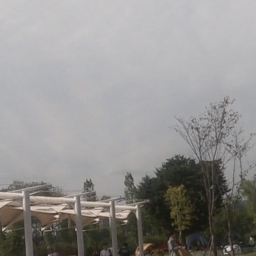} &
            \cellimg{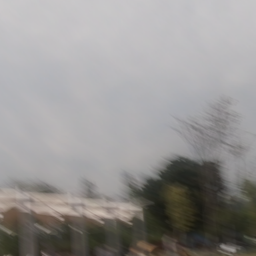} &
            \cellimg{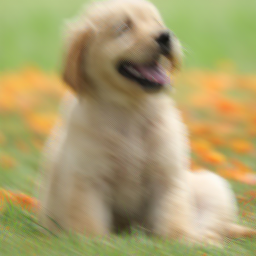}\\
            \cellimg{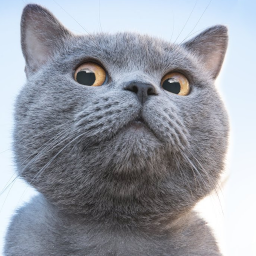} &
            \cellimg{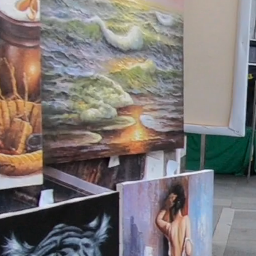} &
            \cellimg{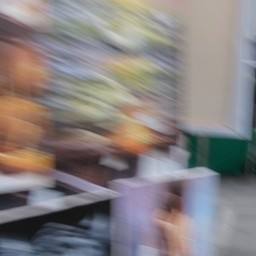} &
            \cellimg{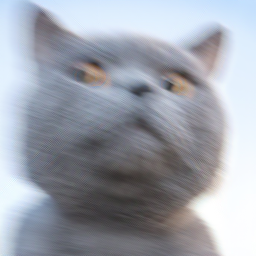}\\
            \cellimg{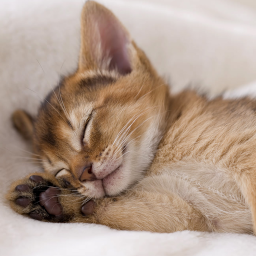} &
            \cellimg{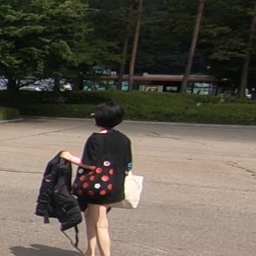} &
            \cellimg{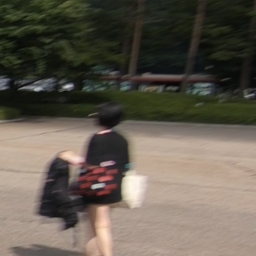} &
            \cellimg{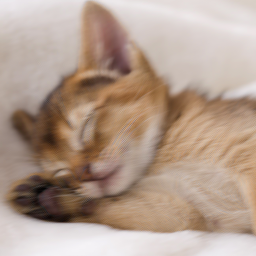}\\
        \end{tabular}
    \caption{\large Transferring blur kernel from the source pair $x, y$ to the target sharp $\hat{x}$ to generate the target blurry image $\hat{y}$}
    \label{fig:synthesis1}
    \end{center}
\end{figure*}

\begin{figure*}
    \huge
    \begin{center}
        \begin{tabular}{cccc}
            $\hat{x}$ & $x$ & $y$ & $\hat{y}$\\
            \cellimg{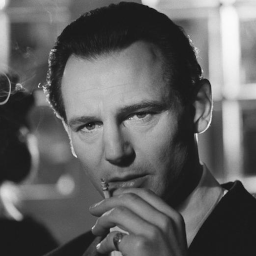} &
            \cellimg{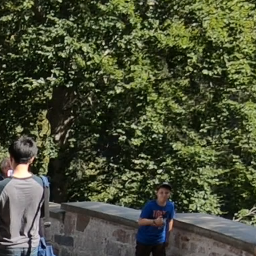} &
            \cellimg{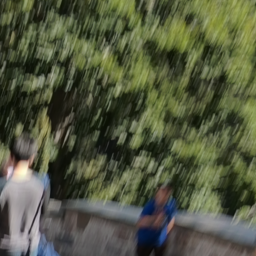} &
            \cellimg{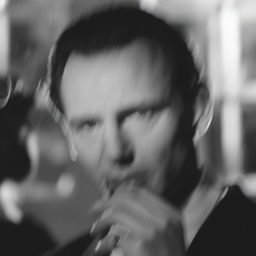}\\
            \cellimg{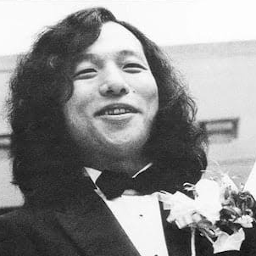} &
            \cellimg{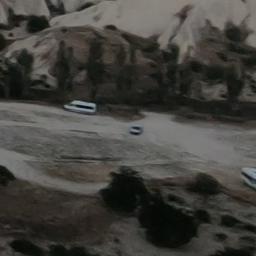} &
            \cellimg{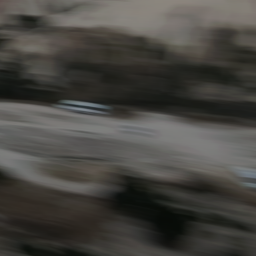} &
            \cellimg{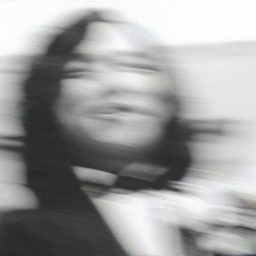}\\
            \cellimg{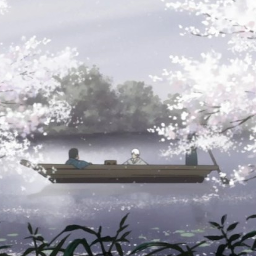} &
            \cellimg{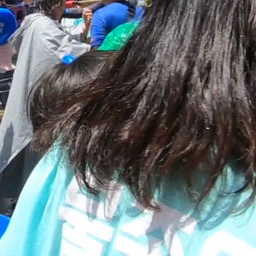} &
            \cellimg{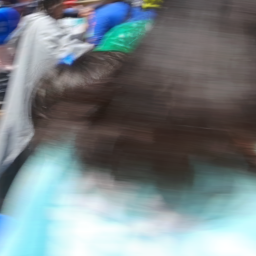} &
            \cellimg{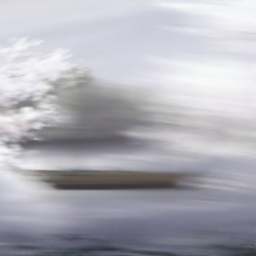}\\
            \cellimg{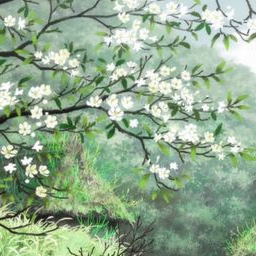} &
            \cellimg{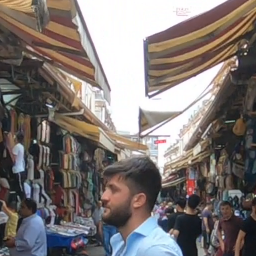} &
            \cellimg{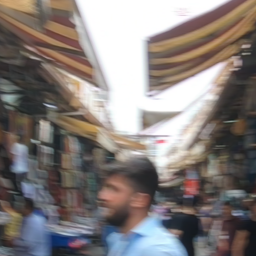} &
            \cellimg{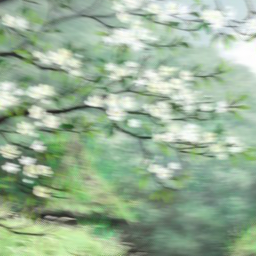}\\
        \end{tabular}
    \caption{\large transferring blur kernel from the source pair $x, y$ to the target sharp $\hat{x}$ to generate the target blurry image $\hat{y}$}
    \label{fig:synthesis2}
    \end{center}
\end{figure*}

Here we provide more qualitative results of our methods including: Blur transferring (\Fref{fig:synthesis1} and \Fref{fig:synthesis2}) and image deblurring on face domain (\Fref{fig:naturaldeblurring1}, \Fref{fig:naturaldeblurring2}, \Fref{fig:naturaldeblurring5}, \Fref{fig:naturaldeblurring3}, \Fref{fig:naturaldeblurring4}, \Fref{fig:naturaldeblurring6}, \Fref{fig:naturaldeblurring7}, and \Fref{fig:naturaldeblurring8}).

\setlength{\tabcolsep}{0pt}

\setlength{\tabcolsep}{0pt}

\begin{figure*}[t]
    \setlength{\tabcolsep}{0.3pt}
    \large
    \begin{center}
    \begin{tabular}{lll}
        \multicolumn{1}{c}{Blur} & 
        \multicolumn{1}{c}{SelfDeblur \cite{ren2020neural}} & 
        \multicolumn{1}{c}{\cite{kupyn2019deblurgan} REDS}\\
        \cellimgthreeperrow{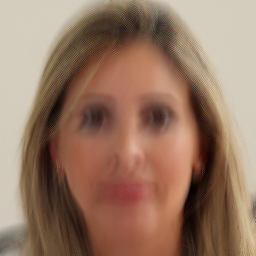} &
        \cellimgthreeperrow{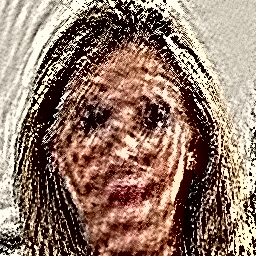} &
        \cellimgthreeperrow{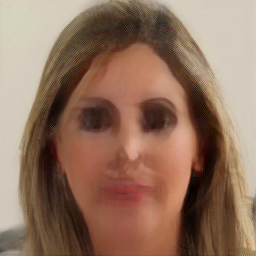}\\[0.2cm]
        \multicolumn{1}{c}{\cite{kupyn2019deblurgan} imgaug} & 
        \multicolumn{1}{c}{\cite{tao2018scale} REDS} & 
        \multicolumn{1}{c}{\cite{tao2018scale} imgaug}\\
        \cellimgthreeperrow{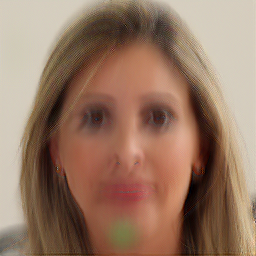} &
        \cellimgthreeperrow{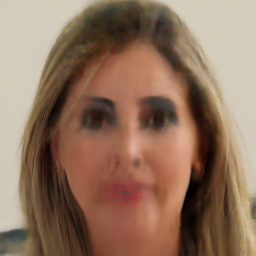} &
        \cellimgthreeperrow{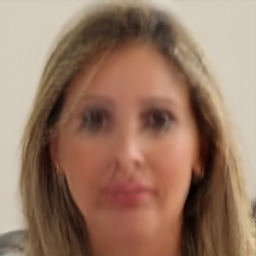}\\[0.2cm]
        \multicolumn{1}{c}{Ours} & &\\
        \cellimgthreeperrow{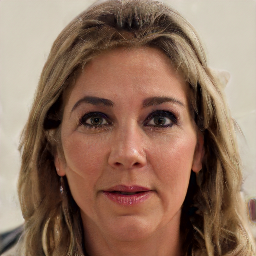} &\\
    \end{tabular}
    \end{center}
    \caption{\large Results of deblurring methods trained on REDS and tested on GOPRO}
    \label{fig:naturaldeblurring1}
\end{figure*}

\begin{figure*}[t]
    \setlength{\tabcolsep}{0.3pt}
    \large
    \begin{center}
    \begin{tabular}{lll}
        \multicolumn{1}{c}{Blur} & 
        \multicolumn{1}{c}{SelfDeblur \cite{ren2020neural}} & 
        \multicolumn{1}{c}{\cite{kupyn2019deblurgan} REDS}\\
        \cellimgthreeperrow{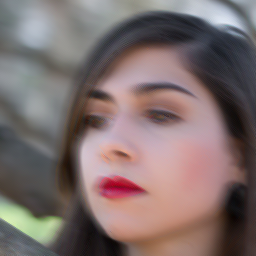} &
        \cellimgthreeperrow{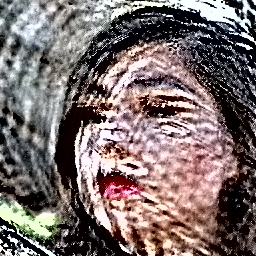} &
        \cellimgthreeperrow{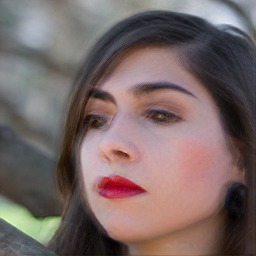}\\[0.2cm]
        \multicolumn{1}{c}{\cite{kupyn2019deblurgan} imgaug} & 
        \multicolumn{1}{c}{\cite{tao2018scale} REDS} & 
        \multicolumn{1}{c}{\cite{tao2018scale} imgaug}\\
        \cellimgthreeperrow{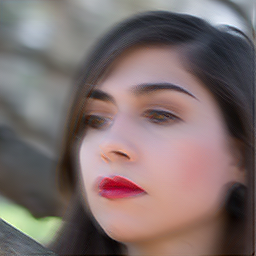} &
        \cellimgthreeperrow{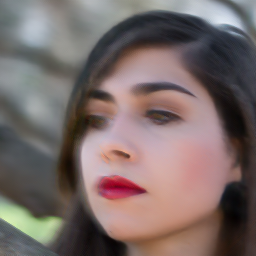} &
        \cellimgthreeperrow{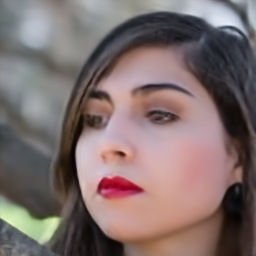}\\[0.2cm]
        \multicolumn{1}{c}{Ours} &  &\\
        \cellimgthreeperrow{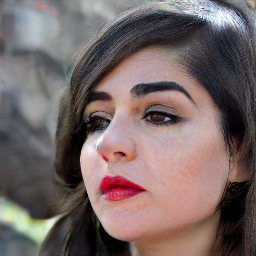} &\\
    \end{tabular}
    \end{center}
    \vskip 0.2in
    \caption{\large Results of deblurring methods trained on REDS and tested on GOPRO}
    \label{fig:naturaldeblurring2}
\end{figure*}

\begin{figure*}[t]
    \setlength{\tabcolsep}{0.3pt}
    \large
    \begin{center}
    \begin{tabular}{lll}
        \multicolumn{1}{c}{Blur} & 
        \multicolumn{1}{c}{SelfDeblur \cite{ren2020neural}} & 
        \multicolumn{1}{c}{\cite{kupyn2019deblurgan} REDS}\\
        \cellimgthreeperrow{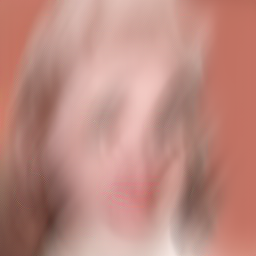} &
        \cellimgthreeperrow{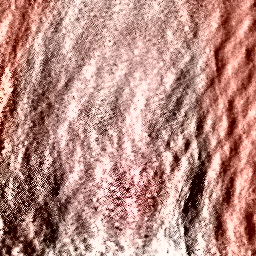} &
        \cellimgthreeperrow{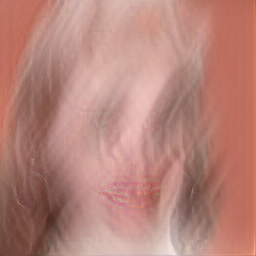}\\[0.2cm]
        \multicolumn{1}{c}{\cite{kupyn2019deblurgan} imgaug} & 
        \multicolumn{1}{c}{\cite{tao2018scale} REDS} & 
        \multicolumn{1}{c}{\cite{tao2018scale} imgaug}\\
        \cellimgthreeperrow{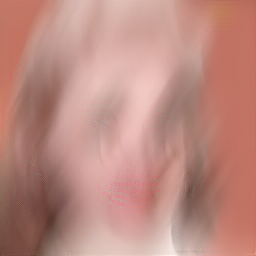} &
        \cellimgthreeperrow{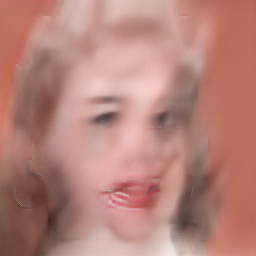} &
        \cellimgthreeperrow{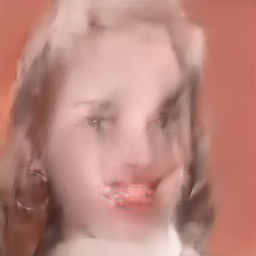}\\[0.2cm]
        \multicolumn{1}{c}{Ours} & &\\
        \cellimgthreeperrow{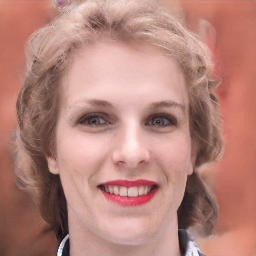} &
    \end{tabular}
    \end{center}
    \vskip 0.2in
    \caption{\large Results of deblurring methods trained on REDS and tested on GOPRO}
    \label{fig:naturaldeblurring5}
\end{figure*}

\begin{figure*}[t]
    \setlength{\tabcolsep}{0.3pt}
    \large
    \begin{center}
    \begin{tabular}{lll}
        \multicolumn{1}{c}{Blur} & 
        \multicolumn{1}{c}{SelfDeblur \cite{ren2020neural}} & 
        \multicolumn{1}{c}{\cite{kupyn2019deblurgan} REDS}\\
        \cellimgthreeperrow{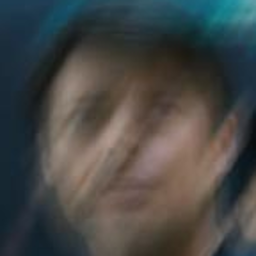} &
        \cellimgthreeperrow{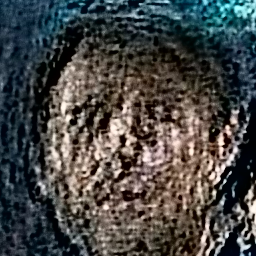} &
        \cellimgthreeperrow{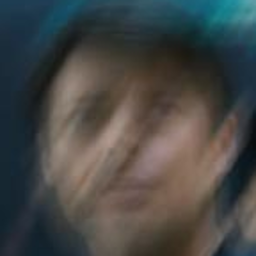}\\[0.2cm]
        \multicolumn{1}{c}{\cite{kupyn2019deblurgan} imgaug} & 
        \multicolumn{1}{c}{\cite{tao2018scale} REDS} & 
        \multicolumn{1}{c}{\cite{tao2018scale} imgaug}\\
        \cellimgthreeperrow{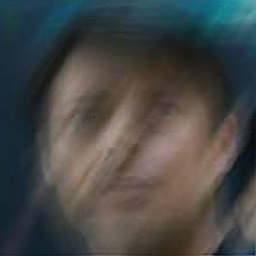} &
        \cellimgthreeperrow{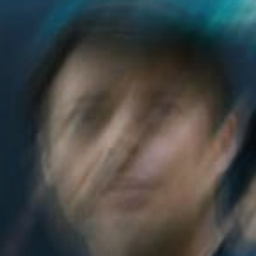} &
        \cellimgthreeperrow{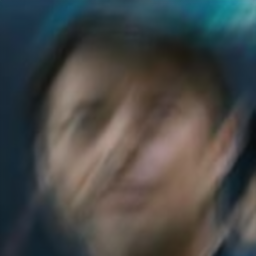}\\[0.2cm]
        \multicolumn{1}{c}{Ours} & &\\
        \cellimgthreeperrow{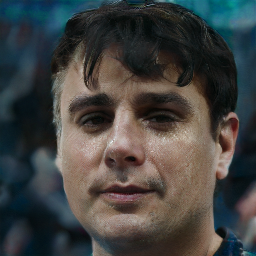} &
    \end{tabular}
    \end{center}
    \vskip 0.2in
    \caption{\large Results of deblurring methods trained on REDS and tested on an in-the-wild example}
    \label{fig:naturaldeblurring3}
\end{figure*}

\begin{figure*}[t]
    \setlength{\tabcolsep}{0.3pt}
    \large
    \begin{center}
    \begin{tabular}{lll}
        \multicolumn{1}{c}{Blur} & 
        \multicolumn{1}{c}{SelfDeblur \cite{ren2020neural}} & 
        \multicolumn{1}{c}{\cite{kupyn2019deblurgan} REDS}\\
        \cellimgthreeperrow{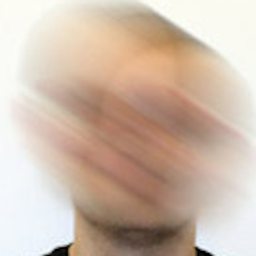} &
        \cellimgthreeperrow{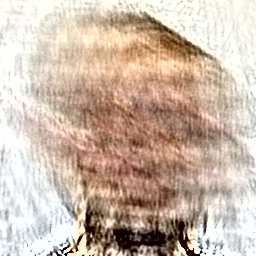} &
        \cellimgthreeperrow{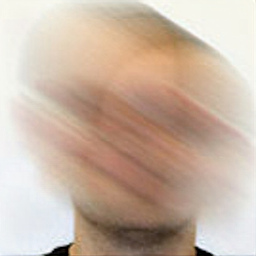}\\[0.2cm]
        \multicolumn{1}{c}{\cite{kupyn2019deblurgan} imgaug} & 
        \multicolumn{1}{c}{\cite{tao2018scale} REDS} & 
        \multicolumn{1}{c}{\cite{tao2018scale} imgaug}\\
        \cellimgthreeperrow{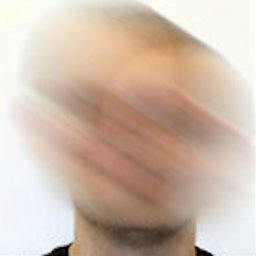} &
        \cellimgthreeperrow{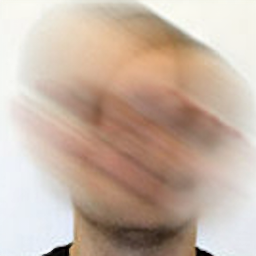} &
        \cellimgthreeperrow{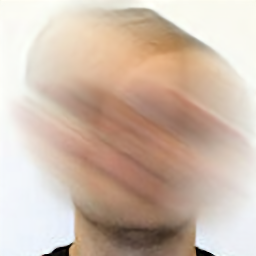}\\[0.2cm]
        \multicolumn{1}{c}{Ours} & &\\
        \cellimgthreeperrow{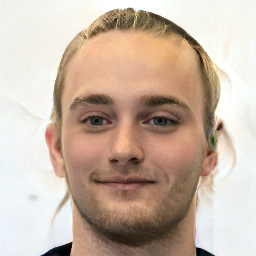} &
    \end{tabular}
    \end{center}
    \vskip 0.2in
    \caption{\large Results of deblurring methods trained on REDS and tested on an in-the-wild example}
    \label{fig:naturaldeblurring4}
\end{figure*}

\begin{figure*}[t]
    \setlength{\tabcolsep}{0.3pt}
    \large
    \begin{center}
    \begin{tabular}{cc}
        \multicolumn{1}{c}{Blur} & 
        \multicolumn{1}{c}{Ours}\\
        \cellbigimg{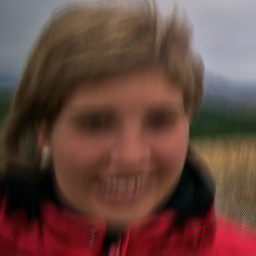} &
        \cellbigimg{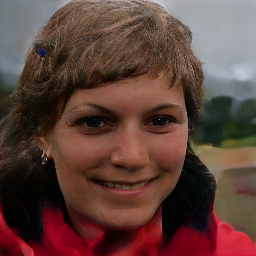}\\
        \cellbigimg{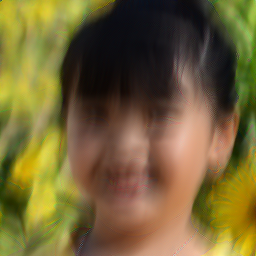} &
        \cellbigimg{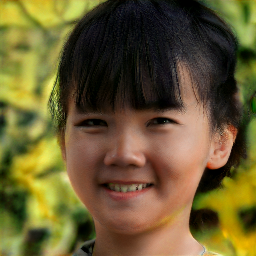}\\
    \end{tabular}
    \end{center}
    \vskip -0.15in
    \caption{\large Results of our method trained on REDS and tested on GOPRO}
    \label{fig:naturaldeblurring6}
\end{figure*}

\begin{figure*}[t]
    \setlength{\tabcolsep}{0.3pt}
    \large
    \begin{center}
    \begin{tabular}{cc}
        \multicolumn{1}{c}{Blur} & 
        \multicolumn{1}{c}{Ours}\\
        \cellbigimg{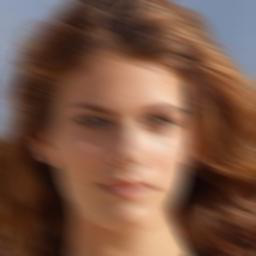} &
        \cellbigimg{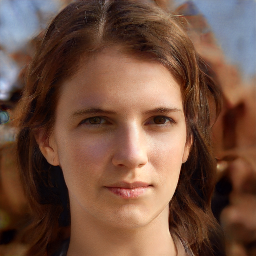}\\
        \cellbigimg{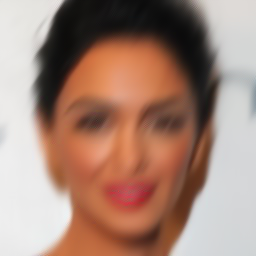} &
        \cellbigimg{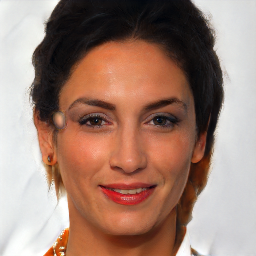}\\
    \end{tabular}
    \end{center}
    \vskip -0.15in
    \caption{\large Results of our method trained on REDS and tested on GOPRO}
    \label{fig:naturaldeblurring7}
\end{figure*}

\begin{figure*}[t]
    \setlength{\tabcolsep}{0.3pt}
    \large
    \begin{center}
    \begin{tabular}{cc}
        \multicolumn{1}{c}{Blur} & 
        \multicolumn{1}{c}{Ours}\\
        \cellbigimg{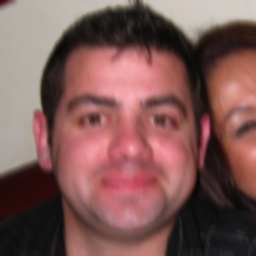} &
        \cellbigimg{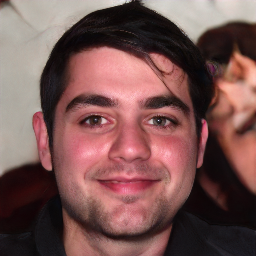}\\
        \cellbigimg{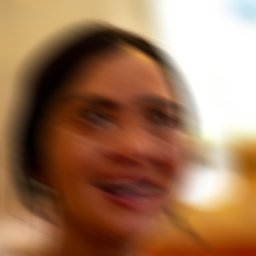} &
        \cellbigimg{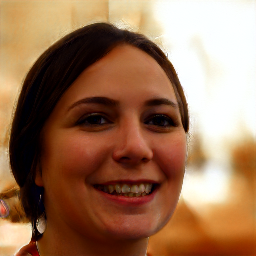}\\
    \end{tabular}
    \end{center}
    \vskip -0.15in
    \caption{\large Results of our method trained on REDS and tested on GOPRO}
    \label{fig:naturaldeblurring8}
\end{figure*}

{\small
\setlength{\bibsep}{0pt}

}